\documentclass{article}

\usepackage{microtype}
\usepackage{graphicx}
\usepackage{subfigure}
\usepackage{booktabs} 

\usepackage{hyperref}

\usepackage[accepted]{arxiv}

\usepackage{amsmath}
\usepackage{amssymb}
\usepackage{mathtools}
\usepackage{amsthm}
\usepackage{multirow}
\usepackage{tabularx}
\usepackage{xcolor}
\usepackage{makecell}
\usepackage{tabularx}
\usepackage{algorithm}
\usepackage{algorithmic}
\usepackage{bm}

\newcolumntype{L}[1]{>{\raggedright\arraybackslash}p{#1}}
\newcolumntype{C}[1]{>{\centering\arraybackslash}p{#1}}
\newcolumntype{R}[1]{>{\raggedleft\arraybackslash}p{#1}}
\newcommand{\tabincell}[2]{\begin{tabular}{@{}#1@{}}#2\end{tabular}}

\usepackage[capitalize,noabbrev]{cleveref}

\theoremstyle{plain}

\theoremstyle{definition}

\theoremstyle{remark}

\usepackage[textsize=tiny]{todonotes}

\icmltitlerunning{Being-0: A Humanoid Robotic Agent with Vision-Language Models and Modular Skills}

\begin{document}

\twocolumn[{
\icmltitle{Being-0: A Humanoid Robotic Agent with Vision-Language Models \\ and Modular Skills}




\icmlaffiliation{pku}{Peking University}
\icmlaffiliation{baai}{BAAI}
\icmlaffiliation{being}{BeingBeyond}


\begin{icmlauthorlist}
\icmlauthor{Haoqi Yuan}{pku,being}
\icmlauthor{Yu Bai}{baai}
\icmlauthor{Yuhui Fu}{pku}
\icmlauthor{Bohan Zhou}{pku}
\icmlauthor{Yicheng Feng}{pku}
\icmlauthor{Xinrun Xu}{baai}
\icmlauthor{Yi Zhan}{pku}
\icmlauthor{Börje F. Karlsson}{baai}
\icmlauthor{Zongqing Lu}{pku,baai,being}
\end{icmlauthorlist}

\icmlcorrespondingauthor{Zongqing Lu}{zongqing.lu@pku.edu.cn}


\vskip 0.1in

\begin{figure}[H]
	\centering
	\hsize=\textwidth
	\includegraphics[width=.99\textwidth, trim={0cm, 5cm, 5cm, 0cm}, clip]{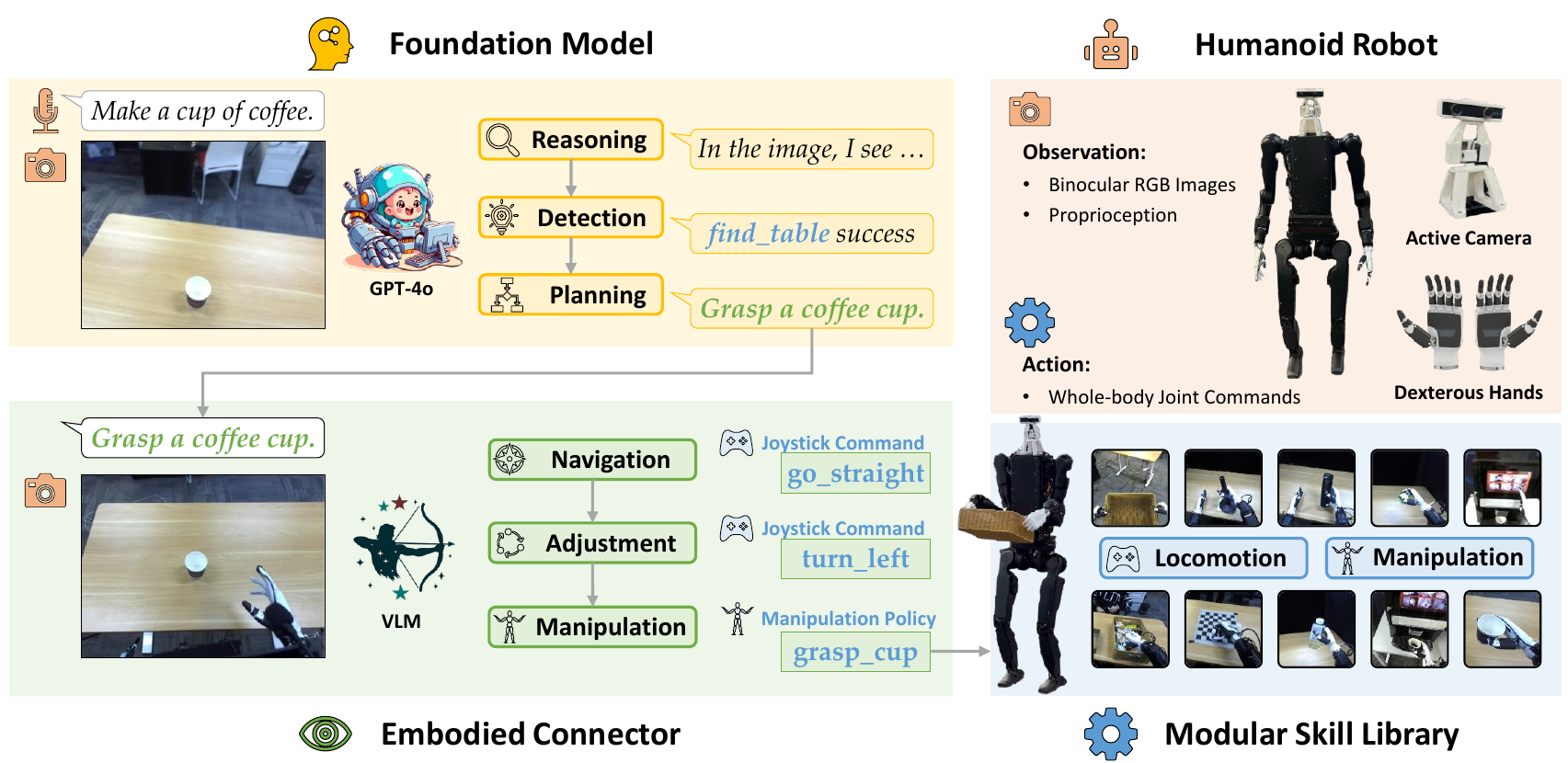}
	\vspace{-2mm}
	\caption{\textbf{Overview of the Being-0 framework.}
		The humanoid agent framework, Being-0, comprises three key components: (1) the Foundation Model (FM) for high-level task planning and reasoning, (2) the Connector, a vision-language model (VLM) that bridges the FM and low-level skills, and (3) the Modular Skill Library for robust locomotion and dexterous manipulation. Together, these components enable Being-0 to effectively control a full-sized humanoid robot equipped with multi-fingered hands and active vision, solving complex, long-horizon embodied tasks in real-world environments.}
	\label{fig:framework}
\end{figure}

\vskip 0.2in

}]



\printAffiliationsAndNotice{}  


\begin{abstract}

Building autonomous robotic agents capable of achieving human-level performance in real-world embodied tasks is an ultimate goal in humanoid robot research. Recent advances have made significant progress in high-level cognition with Foundation Models (FMs) and low-level skill development for humanoid robots. However, directly combining these components often results in poor robustness and efficiency due to compounding errors in long-horizon tasks and the varied latency of different modules.
We introduce \textbf{Being-0}, a hierarchical agent framework that integrates an FM with a modular skill library. The FM handles high-level cognitive tasks such as instruction understanding, task planning, and reasoning, while the skill library provides stable locomotion and dexterous manipulation for low-level control.
To bridge the gap between these levels, we propose a novel \textbf{Connector} module, powered by a lightweight vision-language model (VLM). The Connector enhances the FM’s embodied capabilities by translating language-based plans into actionable skill commands and dynamically coordinating locomotion and manipulation to improve task success.
With all components, except the FM, deployable on low-cost onboard computation devices, Being-0 achieves efficient, real-time performance on a full-sized humanoid robot equipped with dexterous hands and active vision. Extensive experiments in large indoor environments demonstrate Being-0’s effectiveness in solving complex, long-horizon tasks that require challenging navigation and manipulation subtasks. For further details and videos, visit our \href{https://beingbeyond.github.io/Being-0}{project page}.

\end{abstract}

\section{Introduction}
\label{sec:intro}

In the evolving field of embodied AI, humanoid robots represent an ideal platform for achieving human-level intelligence, enabling physical interactions with the real world in ways akin to humans. To realize the ultimate goal of allowing humanoid robots to autonomously perform tasks like humans, current research primarily focuses on improving individual skills, including locomotion \cite{terrain_locomotion, huamnoid_parkour}, bimanual manipulation \cite{iDP3, okami,bidexhd}, and whole-body control \cite{omnih2o, humanplus}. However, building fully autonomous agents for humanoid robots remains a significant and largely unexplored challenge.

An autonomous robotic agent must solve diverse embodied tasks in the real world by grounding language instructions into feasible plans and reliably stitching skills to accomplish long-horizon tasks. Recent studies \cite{roboticsFM1, roboticsFM2} in robotic agents have integrated Foundation Models (FMs) with learning-based robotic skills, leveraging FMs' capabilities in general-purpose vision-language understanding for skill planning \cite{SayCan, VLM-PC}, success detection \cite{inner-monologue}, and reasoning. While these methods have achieved some success in building agents for robot arms \cite{code-as-policy}, wheeled robots \cite{SayCan}, and quadruped robots \cite{VLM-PC}, \textit{can the same success be replicated for humanoid robots?} In this paper, we introduce \textbf{Being-0}, a hierarchical agent framework designed for humanoid robots.

We begin by equipping a universal FM-based agent framework \cite{cradle} with a modular robotic skill library. This skill library includes a robust locomotion skill based on joystick commands and a set of manipulation skills with language descriptions, acquired through state-of-the-art teleoperation \cite{open-television} and imitation learning \cite{ACT} methods. These skills enable the robot to walk and manipulate objects in response to language commands. In principle, the FM agent could call these skills based on image observations in a closed-loop manner to solve long-horizon tasks. However, we find that humanoid robots introduce unique challenges for such a system.

Unlike wheeled robots, which can precisely follow planned navigation trajectories and stop at specific positions for object manipulation, humanoid robots face inherent instability in bipedal locomotion. This instability necessitates frequent adjustments to locomotion commands for error correction. However, existing FMs, such as GPT-4o, suffer from limitations in inference efficiency and embodied scene understanding, making humanoid agents less reactive and robust during the alternating phases of navigation and manipulation in long-horizon tasks.

To address these challenges, we propose a novel \textbf{Connector} module, which serves as an intermediate layer between the FM and skill library in Being-0. The Connector generates real-time commands for both locomotion and manipulation skills based on the FM's language plan and visual observations. We model the Connector as a vision-language model (VLM) and train it using first-person images of indoor navigation annotated with language instructions, object labels, and bounding boxes. This training scheme distills embodied knowledge from vision-language navigation data into the lightweight VLM-based Connector, enabling accurate skill planning and efficient navigation at a higher control frequency. Furthermore, to seamlessly stitch navigation and manipulation skills, the Connector can send locomotion commands to adjust the humanoid's pose, improving the initialization state for subsequent manipulation tasks.

We conduct extensive experiments on navigation, manipulation, and long-horizon tasks using a full-sized humanoid robot equipped with dexterous hands and an active camera. The results demonstrate that Being-0 achieves an average completion rate of 84.4\% on challenging long-horizon tasks, highlighting the significant contribution of the Connector module and the use of active vision in the system. By deploying all modules -- except for the FM on the cloud -- on onboard computation devices, Being-0 achieves 4.2$\times$ efficiency in navigation compared to fully FM-based agents.

Our contributions can be summarized as follows:
\begin{itemize}
\vspace{-3mm}
\setlength{\itemsep}{0em}
\item We propose a hierarchical agent framework for humanoid robots, where each layer is optimally deployed on either the cloud or onboard devices, enabling efficient execution of long-horizon embodied tasks.
\item We introduce a VLM-based Connector module to bridge the gap between the FM's language-based task plans and the execution of low-level skills. This module enhances embodied decision-making and effectively coordinates locomotion and manipulation skills for humanoid robots.
\item Our agent is capable of controlling humanoid robots with multi-fingered dexterous hands and active cameras, enhancing their dexterity in both navigation and manipulation tasks.
\vspace{-3mm}
\end{itemize}

\begin{figure*}[!t]
\centering
\includegraphics[width=.95\linewidth, trim={0.5cm, 7cm, 2cm, 0cm}, clip]{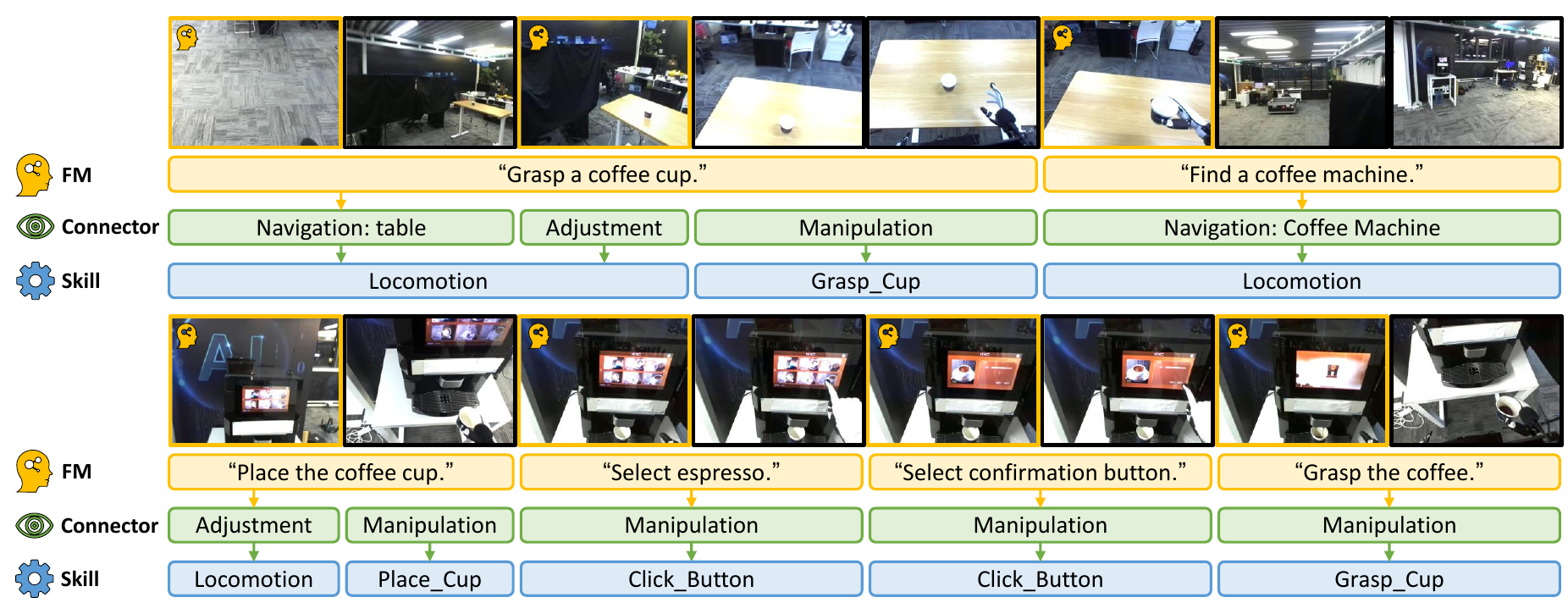}
\caption{\textbf{Workflow of Being-0 for the task ``make a cup of coffee".} The figure illustrates the step-by-step execution of the task, with images arranged in two rows. The execution order proceeds left to right in the first row, then continues left to right in the second row. Images with yellow borders indicate decision-making points for the Foundation Model (FM). The yellow dialog boxes display the FM's plans, the green boxes show decisions made by the Connector, and the blue boxes represent the skills called from the modular skill library. }
\label{fig:workingflow}
\end{figure*}

\section{Humanoid Robot and Agent}

As illustrated in Figure \ref{fig:framework}, we consider a humanoid robot with 41 degrees of freedom (DoFs), including a 13-DoF lower body (two legs and a torso), two 7-DoF arms, two 6-DoF dexterous hands, and a 2-DoF neck. The multi-fingered dexterous hands enable complex, human-like manipulation, while the actuated neck, equipped with a binocular RGB camera, provides active vision. This hardware configuration grants the robot human-level dexterity in visual perception, navigation, and object interaction.

An autonomous agent aims to complete real-world tasks described in natural language by controlling the robot's whole-body joints. Formally, at any time, the agent has access to a task description $l$ (e.g. ``make a cup of coffee") and can query the robot's observations, including:
(1) \textbf{Proprioception}: $\left( q^l, q^u, q^h; \dot{q}; \omega \right)$, where $q^l, q^u, q^h$ denote joint positions of the lower body, upper body, and neck, respectively; $\dot{q}$ represents joint velocity; and $\omega$ is the root velocity and angular velocity acquired from the IMU. 
(2) \textbf{Visual input}: binocular RGB images $o^l, o^r$ from the left and right cameras. 
The agent can take actions $(a^l, a^u, a^h)$, which specify target joint positions for the PD controller of the lower body, upper body, and neck, respectively.

Instead of directly mapping task descriptions and observations into muscle actuation, humans rely on a hierarchical system to solve real-world tasks. For example, the task ``make coffee" is first decomposed into detailed plans such as ``find a cup, grasp a cup, find a coffee machine..." based on prior experience. Then, practiced motor skills, such as walking and grasping, are reused to sequentially execute the task. Recent advances in robotic agents \cite{SayCan,inner-monologue} adopt this approach by integrating a high-level planner with a low-level skill library. In this paper, we aim to build such an agent for humanoid robots, addressing the unique and largely unexplored challenges in this domain.

\section{The Hierarchical Agent Framework}

\subsection{Modular Skill Library}

The first challenge we address is: how can we acquire diverse, robust low-level skills for a humanoid robot to support solving real-world, long-horizon tasks? In the literature on whole-body control \cite{humanplus, omnih2o}, policies for individual skills typically map observations to whole-body target joint positions, simultaneously controlling leg motion and arm manipulation. However, these methods have not yet developed a wide range of manipulation skills due to the complexity of achieving precise manipulation, stable locomotion, and sim-to-real deployment with one policy.

For most tasks, we observe that the lower body and upper body serve distinct functionalities: the lower body is primarily used for navigation, while the upper body is used for manipulation \cite{exbody}. This observation motivates us to develop separate skills for stable lower-body locomotion and upper-body manipulation, building on recent advances \cite{OpenVLA, open-television} that have demonstrated the feasibility of acquiring abundant upper-body manipulation skills while keeping the lower body fixed.

\textbf{Stable Locomotion with Joystick Commands.} The locomotion skill, which controls the lower-body joints, must enable navigation in various directions and maintain stable standing during manipulation tasks. We adopt reinforcement learning (RL) approaches \cite{learning_survey} to train a goal-conditioned proprioceptive policy $\pi^L(a^l|q^l,q^u,\dot{q},\omega; v^g)$ in simulation \cite{isaacgym}, followed by sim-to-real deployment at a control frequency of 50 Hz. Here, $v^g$  represents the joystick velocity command. By incorporating domain randomization and external forces during simulation, this skill enables the robot to walk according to joystick commands while maintaining balance.
To integrate this into the skill library, we define a set of locomotion skills based on different joystick commands, along with skills for adjusting the head for active vision: \{no\_action, go\_straight, walk\_backwards, turn\_left, turn\_right, sidestep\_left, sidestep\_right, tilt\_head, turn\_head\}.

\textbf{Acquiring Manipulation Skills.}
Teleoperation and imitation learning have emerged as promising approaches for acquiring diverse robotic manipulation skills at low cost. To collect high-quality, human-like manipulation data for the humanoid equipped with two dexterous hands and active vision, we use Apple VisionPro for teleoperation, following recent work \cite{open-television}. Binocular image observations $o^l, o^r$ are projected to the VisionPro, and the captured human motions of the head, wrists, and fingers are retargeted to robot actions at a control frequency of 10 Hz.
For each skill, teleoperation trajectories $\tau=\{(o^l_t,o^r_t,q^u_t,q^h_t,a^u_t,a^h_t)\}_{t=1}^T$ are recorded, including robot observations and actions (excluding the lower body). We use ACT~\cite{ACT}, a behavior-cloning method with a Transformer architecture, to train the policy $\pi^{M_i}\left([a^u_j,a^h_j]_{j=t}^{t+K}|o^l_t,o^r_t,q^u_t,q^h_t\right)$ for each manipulation skill $M_i$, associated with a language description such as ``grasp\_bottle". The length of the predicted action sequence, $K$, is set to 30 during training and 10 during deployment. 
This approach ensures scalability of the skill library, as a new skill can be acquired with $50\sim 150$ trajectories, requiring less than 1 hour of teleoperation.

\subsection{Foundation Model}
The high-level planner of the agent makes skill-level decisions across diverse tasks and environments, necessitating strong capabilities in general-purpose vision-language understanding and reasoning. Foundation Models (FMs) excel in these areas and have been widely adopted in recent research on AI agents \cite{llm-agent-survey, cradle}. For example, Cradle \cite{cradle}, an agent framework built on GPT-4o, has been successfully applied to open-world games and software usage, operating keyboard and mouse skills based on image observations. Inspired by this work, we adapt the Cradle framework to build a generalist agent for humanoid robots, enabling the robot to operate skills from the skill library and solve real-world tasks.

Given an instruction $l$ and an image observation $o^l$, the FM (GPT-4o) performs three key functionalities for decision-making:
(1) \textbf{Reasoning}: The FM generates a description of the observed image and instruction, aiding in task understanding and identifying the current stage of execution. 
(2) \textbf{Detection}: The FM evaluates the success of recently executed skills, identifying failures and exceptions to inform task planning.
(3) \textbf{Planning}: Based on the reasoning and detection results, the FM selects the next skill to execute from the skill library.
Detailed prompt designs for the FM can be found in Appendix~\ref{app:prompt}.

However, when directly integrating the FM with the skill library, we encounter several challenges that severely hinder system performance. The inherent instability of bipedal locomotion makes the humanoid's position unpredictable after short periods of walking, necessitating frequent adjustments to joystick commands rather than executing open-loop command sequences. Additionally, existing FMs, including GPT-4o, struggle with accurate 3D scene understanding, often failing to estimate the direction and depth of navigation targets correctly, which can lead to incorrect skill plans (see experimental results in Figure \ref{fig:app-wo-connector}). Even when the agent successfully navigates to a target location (e.g., a table), its final standing position may not provide a suitable initial state for subsequent manipulation skills (e.g., ``grasp\_cup"), resulting in task failure (see Figure \ref{fig:app-wo-adjust}). Furthermore, the low inference speed of large FMs significantly reduces system efficiency, causing the robot to move slowly and react less promptly to dynamic environments.

To address these challenges, we propose a novel Connector module in Being-0, which bridges the gap between the FM and the skill library, enhancing real-time, embodied decision-making.


\section{Embodied Connector}

The primary goal of the Connector is to translate high-level language-based plans generated by the FM into executable skill commands reliably and efficiently. At the core of the Connector is a lightweight vision-language model (VLM) trained on annotated navigation data, which enhances the agent’s embodied capabilities. This VLM enables several downstream functionalities, including grounded skill planning, closed-loop navigation, and improved transitions between navigation and manipulation during long-horizon task execution.

\subsection{Training the Vision-Language Model}

To equip the VLM with spatial and object understanding, as well as the ability to anticipate future skills based on context, we train it on a dataset of first-person navigation images. These images are annotated with language descriptions, skills, object labels, and bounding boxes. We adopt VideoLLaMA2~\citep{videollama2} as the backbone architecture, using image observations and text instructions as inputs. The model is optimized through multi-task learning, encompassing image description, skill prediction, and object detection. The trained VLM achieves an average inference time of approximately 1 second on onboard devices across all tasks, significantly outperforming the latency of GPT-4o on cloud services. Further details on the dataset and training process are provided in Appendix~\ref{app:connector}.

\subsection{Grounded Skill Planning}

The main usage of the VLM is to convert the FM’s language-based plans and real-time image observations into actionable skill plans, such as navigation targets or manipulation skills. By leveraging its enhanced understanding of relative 3D object locations, the VLM not only grounds the FM's plans into executable skills but also corrects or refines them when necessary. For example: If the FM generates a plan to ``grasp a cup" but the robot is still far from the table, the VLM interprets ``grasp a cup" as a long-term goal and outputs the feasible skill (e.g., ``move\_towards(table).'').
Conversely, if the FM plans to ``find a table" but the robot is already at the table, the VLM’s navigation functionality (Section~\ref{sec:composite_skills}) signals success to the FM, prompting it to proceed to the next skill through reasoning. 
This capability ensures that the planned skills remain grounded in the physical environment, reducing errors and improving task success rates.

\subsection{Visual Navigation with Locomotion Skills}
\label{sec:composite_skills}

To enable the robot to reach visual navigation goals (e.g., a table), the Connector leverages the VLM's visual understanding and object detection capabilities. When the goal object is within the robot’s field of view, the Connector estimates its relative position using the detected bounding box and synthetic depth from binocular images. Based on this estimation, the VLM selects the most appropriate locomotion skill to move towards the object's direction. If the object is not visible, the VLM triggers an exploration routine, combining locomotion skills with active camera movements to search for the goal. This approach significantly enhances the robot’s ability to locate objects compared to systems with fixed cameras. Implementation details are provided in Appendix~\ref{app:connector}.
By integrating the VLM’s efficient inference capabilities with modular locomotion skills, this method accelerates humanoid robot navigation while maintaining robustness in dynamic environments.


\subsection{Coordinating Navigation and Manipulation}
To address the challenge that navigation processes may terminate in suboptimal poses for subsequent manipulation skills, we propose a pose adjustment method using the VLM. During navigation, the VLM predicts not only the object’s bounding box but also the optimal alignment direction for the robot relative to the object. If the robot’s current facing direction deviates from this alignment, the VLM triggers a composite skill combining head rotation and forward movement to adjust the robot’s pose. This allows the robot to approach the target object along an arc-shaped path, ensuring it reaches an optimal position for manipulation. Further details are provided in Appendix~\ref{app:connector}.

\subsection{Summary}
Figure \ref{fig:workingflow} illustrates the workflow of Being-0, highlighting the role of the Connector module.
In summary, the embodied Connector provides several critical advantages for executing long-horizon tasks. By leveraging the lightweight VLM, the Connector ensures real-time responsiveness, enabling the robot to adapt dynamically to changes in its environment. This real-time capability is essential for efficient task execution, as the Connector dynamically selects and sequences modular skills, significantly reducing operational latency. 
Unlike the FM, the VLM’s enhanced spatial understanding allows the robot to accurately perceive and respond to its surroundings, grounding abstract language-based plans in real-time visual input. This spatial reasoning capability is particularly valuable in complex tasks, where the Connector’s robustness ensures adaptability to unexpected obstacles or environmental variations. 
Additionally, the Connector facilitates improved transitions between navigation and manipulation by adjusting the robot’s pose, ensuring that the robot reaches the proper positions for subsequent skills. 
Together, these features make the embodied Connector a cornerstone of Being-0, enabling it to tackle challenging, long-horizon tasks that require both navigation and manipulation in complex environments.


\section{Experiments}
\subsection{Real-World Setup}

We conduct experiments on a Unitree H1-2 humanoid robot equipped with two Inspire hands for manipulation, two Dynamixel motors for neck movement, and a ZED-mini camera mounted on the neck for active vision. The NVIDIA Jetson AGX onboard device is used to deploy the Connector and all modular skills.

Our experimental environment is a large office scene spanning a 20$\times$20 (m) area, featuring multiple office cubicles, a wooden table, a coffee machine, and corridors connecting reception and meeting rooms. This complex and richly populated environment provides a challenging benchmark for evaluating navigation and long-horizon task execution capabilities.

To build the manipulation skill library, we collect data for a variety of daily manipulation tasks, including single-hand and bimanual tasks such as grasping and placing objects, operating a basket with items, using a coffee machine, and playing with toy bricks and chess games. The data collection and training details are presented in Appendix~\ref{app:manip-skills}.

We evaluate the agent on a diverse set of long-horizon tasks designed to test the system’s robustness in task planning and skill execution. These tasks include:
\begin{itemize}
\vspace{-3mm}
\setlength{\itemsep}{0em}
    \item \textbf{Fetch-bottle} and \textbf{Deliver-basket}: These tasks require the robot to navigate to a distant wooden table and perform manipulation tasks.
    \item \textbf{Prepare-coffee}, \textbf{Make-coffee}, and \textbf{Deliver-coffee}: These are particularly challenging tasks composed of multiple subtasks, including precise manipulation skills such as pressing buttons on the coffee machine and placing a cup in the correct position. 
\vspace{-3mm}
\end{itemize}
Further details on the task setups are provided in Appendix \ref{app:exp-detail}.

\renewcommand\tabcolsep{16pt}

\begin{table}[!t]
\renewcommand\arraystretch{1.2}
\caption{Task completion rates for Being-0 with and without the Connector across various long-horizon tasks. The results demonstrate significant performance improvements when the Connector is used. }
\label{table:main}
\vspace{0.2cm}
\centering
\small
\begin{tabular}{p{2.2cm}p{1.3cm}p{1.3cm}} 
\toprule
\textbf{Task} & \textbf{\tabincell{c}{{w/o Connector}}}  &  \textbf{\tabincell{c}{{Being-0}}} \\ \midrule
\textbf{Fetch-bottle} & 0.00 & \textbf{0.90} \\ 
\textbf{Deliver-basket}  &  0.00 & \textbf{0.80} \\ 
\textbf{Prepare-coffee}   & 0.00 & \textbf{0.75} \\ 
\textbf{Make-coffee}   & \textbf{0.90} & \textbf{0.90} \\ 
\textbf{Deliver-coffee}   & 0.33 & \textbf{0.87} \\ 
\bottomrule 
\end{tabular}
\end{table}

\subsection{Solving Long-Horizon Embodied Tasks}
We evaluate the performance of Being-0 on long-horizon embodied tasks, with the main results presented in Table~\ref{table:main}. These tasks are designed to test the robot's ability to execute complex sequences of skills in real-world environments, requiring precise coordination between high-level cognition and low-level skills.

The results demonstrate a significant performance improvement when the Connector module is utilized, particularly for tasks requiring multiple steps and integration of different skills. For example, in the Fetch-bottle task, the baseline system (w/o Connector) achieves a score of 0.00, whereas the system with the Connector attains a remarkable score of 0.90. Similarly, tasks such as Deliver-basket and Prepare-coffee show substantial improvements, with performance increasing from 0.00 to 0.80 and 0.00 to 0.75, respectively.

These findings highlight the critical role of the Connector in enabling the robot to effectively complete long-horizon tasks. By bridging the gap between the FM and the skill library, the Connector enhances task success rates, particularly for scenarios requiring complex, sequential skills. Overall, the results confirm that Being-0 is highly capable of executing long-horizon tasks with robust and reliable performance.

\begin{table}[!t]
\renewcommand\arraystretch{1.2}
\caption{Ablation study on the proposed adjustment method in the Connector module. The results indicate the number of successful manipulations out of 5 navigation trials. \textbf{\textit{(t)}} denotes ``on the table" and \textbf{\textit{(m)}} denotes ``on the coffee machine".  }
\label{table:adjust}
\vspace{0.2cm}
\centering
\small
\begin{tabular}{p{2.4cm}p{1.2cm}p{1.2cm}} 
\toprule
\textbf{Task} & \textbf{\tabincell{c}{{w/o Adjust.}}}  &  \textbf{\tabincell{c}{{Being-0}}} \\ \midrule
\textbf{Grasp-bottle} & 2 / 5 & \textbf{4} / 5 \\ 
\textbf{Place-basket}  & \textbf{4} / 5 & 3 / 5 \\ 
\textbf{Grasp-coffee}   & 1 / 5 & \textbf{4} / 5 \\ 
\textbf{Place-coffee \textit{(t)}}   & 4 / 5 & \textbf{5} / 5 \\ 
\textbf{Place-coffee \textit{(m)}}   & 0 / 5 & \textbf{3} / 5 \\ 
\bottomrule 
\end{tabular}
\end{table}

\begin{table*}[!t]
\renewcommand\arraystretch{1.2}
\caption{Success rates of navigation and manipulation tasks with different active camera configurations. The number following \textbf{Fixed Cam.} denotes the pitch angle set for the camera in the absence of active neck movement.}
\label{table:active_camera}
\vspace{0.2cm}
\centering
\small
\begin{tabular}{lccccc}
\toprule
\multicolumn{1}{l}{\multirow{2}{*}{\textbf{Method}}} & \multicolumn{2}{c}{\textbf{Navigation}}  & \multicolumn{2}{c}{\textbf{Manipulation}}  \\ \cmidrule(lr){2-5} 
\multicolumn{1}{c}{}  & table & coffee machine  & grasp coffee & place coffee \\ \midrule

\textbf{\tabincell{c}{{Fixed Cam. (0.3)}}}   &  \textbf{5} / 5 & \textbf{5} / 5 & 0 / 5 & 0 / 5 \\ 

\textbf{\tabincell{c}{{Fixed Cam. (0.6)}}}   & 0 / 5 & 0 / 5 & 2 / 5 & 1 / 5 \\ 

\textbf{\tabincell{c}{{Fixed Cam. (0.9)}}}   & 0 / 5 & 0 / 5 & 4 / 5 & \textbf{5} / 5 \\ 

\textbf{Being-0 (Active Cam.) }  & \textbf{5} / 5 & \textbf{5} / 5 & \textbf{5} / 5 & \textbf{5} / 5 \\ 
\bottomrule 
\end{tabular}
\end{table*}

\subsection{Ablation Study}
\textbf{Adjustment in Navigation. }
We evaluate the proposed adjustment method in the Connector by testing the agent on two-stage tasks that involve navigation followed by manipulation. In this setup, the success rate of the manipulation task directly reflects the quality of the robot’s termination state after navigation. Table~\ref{table:adjust} presents the results comparing Being-0 with and without adjustment.

For grasping tasks, such as Grasp-bottle and Grasp-coffee, Being-0 with adjustment significantly outperforms the ablation baseline, achieving success rate gains of over 0.4. This improvement can be attributed to the robot’s ability to terminate navigation in positions that are favorable for grasping. Without adjustment, the robot may stop too far from the object or position the object behind the grasping hand, causing the subsequent grasping skill to fail (see Figure \ref{fig:app-wo-adjust}).

Placing tasks on the table, including Place-basket and Place-coffee \textit{(t)}, are less sensitive to adjustment. This is because, as long as the robot reaches the table, it can successfully place the object, regardless of its standing pose relative to the table. However, for Place-coffee \textit{(m)}, which requires placing the cup on a coffee machine with a very small available area, Being-0 with adjustment performs significantly better.

These results demonstrate that the proposed adjustment method enhances performance in sequential navigation and manipulation tasks, particularly for manipulation tasks where the success depends heavily on the robot’s initial state relative to the object.

\textbf{Active Vision. }
The active camera is a core hardware component of our system, significantly enhancing the robot’s dexterity across various skills. We conduct an ablation study to evaluate the performance of Being-0 when using a fixed camera with different pitch angles, compared with the active camera configuration. Given that the camera’s pitch angle impacts both navigation and manipulation performance, we test the agent with fixed camera setups at various angles. Table~\ref{table:active_camera} presents the results across different tasks.

For navigation tasks, we observe that a small pitch angle (Fixed Cam. (0.3)) yields good performance, while larger pitch angles result in failure. This is because a camera with a large pitch primarily views the ground, causing the agent to lose sight of navigation targets. In contrast, for tabletop manipulation tasks, higher pitch angles improve success rates, as the robot needs to look downward to locate objects on the table.

However, no fixed camera configuration achieves high success rates for both navigation and manipulation tasks. In comparison, Being-0 with an active camera consistently achieves perfect success rates across all tasks. These results underscore the significant advantage of an active camera, enabling the robot to dynamically adapt its field of view to meet the requirements of diverse tasks.

\textbf{Efficiency. }
Being-0 demonstrates notable advantages in efficiency, primarily due to the inclusion of the proposed Connector module. To evaluate this, we conduct an ablation study on the task ``navigate to the wooden table," with the results presented in Table~\ref{table:nav_efficiency}.

The results indicate that Being-0 with the Connector achieves a 4.2$\times$ increase in moving speed compared to the configuration without the Connector, along with a perfect success rate of 5/5. In contrast, the agent without the Connector consistently fails to reach the distant target. This is because GPT-4o alone frequently makes errors in planning locomotion directions, leading to inefficient or incorrect navigation paths.
These findings highlight the critical role of the Connector module in enhancing the efficiency of the Being-0 framework.

\begin{table}[!t]
\renewcommand\arraystretch{1.2}
\caption{Ablation study on the efficiency of Being-0 in navigation. The table reports the average moving speed (cm/s) and success rates for various agent configurations.}
\label{table:nav_efficiency}
\vspace{0.2cm}
\centering
\small
\resizebox{0.99\linewidth}{!}{
\begin{tabular}{lcc}
\toprule
\textbf{Method} &  \textbf{Avg. Speed} & \textbf{Success}  \\ \midrule

\textbf{\tabincell{c}{{w/o Connector}}}  & 2.3 &  0 / 5 \\ 

\textbf{\tabincell{c}{{Fixed Cam. (0.3)}}}  & 8.5 & \textbf{5} / 5  \\ 

\textbf{\tabincell{c}{{Being-0}}}   & \textbf{9.6} &  \textbf{5} / 5 \\ 
\bottomrule 
\end{tabular}}
\end{table}

\subsection{Robustness and Scalability}
\textbf{Navigation. }
To assess the robustness of Being-0 in navigation, we test it across various scene configurations and tasks. The results, shown in Table~\ref{table:navigation}, demonstrate that Being-0 consistently achieves high success rates across all settings.

For navigation to targets within the same room, Being-0 achieves a perfect success rate of 1.0. When adapting to unseen layouts with obstacles, it maintains strong performance with a slight drop of 0.2 in success rate. Additionally, Being-0 successfully handles cross-room navigation tasks, achieving a high success rate of 0.83. These tasks require multi-step reasoning and planning by the FM. For example, to locate the reception table, the robot must first identify and navigate to the room’s exit before proceeding further.

\begin{table}[!t]
\renewcommand\arraystretch{1.2}
\caption{Navigation performance across various scene configurations and target locations.  }
\label{table:navigation}
\vspace{0.2cm}
\centering
\small
\begin{tabular}{lc}
\toprule
\textbf{Task} &  \textbf{Success}  \\ \midrule

\textbf{In-room} & 1.00 \\

\textbf{In-room with obstacles} &  0.80 \\ 

\textbf{Cross-room}  &  0.83 \\ 
\bottomrule 
\end{tabular}
\end{table}

\begin{table}[!t]
\renewcommand\arraystretch{1.2}
\caption{Performance of manipulation skills across different scene configurations, including seen objects, unseen objects, and scenarios with visual perturbations. \textbf{*} denotes the use of dexterous hands equipped with tactile sensors.}
\label{table:manip}
\vspace{0.2cm}
\centering
\small
\begin{tabular}{@{}p{2.4cm}p{0.5cm}p{0.5cm}p{1.3cm}@{}} 
\toprule
\textbf{Task} &  \textbf{Seen Obj.} & \textbf{Unseen Obj.} & \textbf{Perturb.} \\ \midrule

\textbf{Grasp-bottle}   & 0.86 & 0.63 & 0.77  \\ 

\textbf{Handout-snack}  & 0.90 & 1.00 & 0.80  \\ 

\textbf{Place-pole}  & 0.90 & \--\-- & 0.80  \\ 

\textbf{Play-chess*} & 0.90 & \--\-- & 0.90 \\ 
\bottomrule 
\end{tabular}
\end{table}

\textbf{Manipulation Skills. }
Table~\ref{table:manip} presents the performance of manipulation skills across various settings. The success rate shows a slight decline when handling unseen objects or encountering visual perturbations, demonstrating the robustness and generalizability of the learned manipulation policies.

Furthermore, the same framework used for acquiring manipulation skills can be extended to dexterous hands equipped with tactile sensors (the Play-chess task in Table~\ref{table:manip}) and tasks requiring precise manipulation of small objects (see Figure \ref{fig:app-manipulation}), demonstrating the scalability of the skill library to support more complex and challenging manipulation tasks.

\section{Related Work}

\textbf{Humanoid Locomotion and Manipulation.} 
Humanoid robots \cite{humanoid_robot, humanoid-locomotion-and-manipulation} are considered an ideal morphology for human-designed environments, where locomotion and manipulation are fundamental skills. Early approaches focused on locomotion using optimal control \cite{dynamic_walk, whole_body_humaniod_control, optimal_control}, while recent advances have successfully trained locomotion policies with reinforcement learning (RL) and sim-to-real techniques \cite{learning_survey}, achieving robust walking on flat ground \cite{cassie_locomotion}, complex terrains \cite{rl_upstair, terrain_locomotion, rl_locomotion}, and advanced parkour skills \cite{huamnoid_parkour}.
For manipulation, while RL-based methods \cite{crossdex,resdex} suffer from the significant sim-to-real gap, imitation learning with teleoperation data has been a dominant approach due to its simplicity and effectiveness. Research has explored diverse teleoperation schemes, leveraging VR \cite{open-television}, exoskeletons \cite{mobile_aloha, ace}, or cameras \cite{humanplus}. Improved imitation learning methods, such as Diffusion Policies \cite{diffusion-policy, 3D-diffusion-policy, iDP3} and ACT \cite{ACT}, have further advanced training performance.
Recently, whole-body control \cite{humanplus, omnih2o, hover, exbody2} has gained attention for integrating locomotion and manipulation within a single policy. However, this remains challenging due to the combined complexities of both fields.

\textbf{Embodied Agents} \cite{roboticsFM1,roboticsFM2} for robotics require not only low-level skills but also strong capabilities in common-sense reasoning for high-level decision-making. Recent research has explored two primary approaches to building embodied agents with Foundation Models (FMs).
The first approach directly applies existing FMs, pre-trained on Internet-scale datasets, to robotic tasks without additional fine-tuning. These models leverage their strong general-purpose vision-language understanding capabilities for embodied tasks such as planning \cite{SayCan, plan4mc, VLM-PC, smart-llm} and reasoning \cite{inner-monologue, creative-agents, rl-gpt}. These methods typically rely on a predefined skill library for low-level execution.
The second approach focuses on training robotic FMs using extensive robotic datasets. Notable efforts include Robotic Transformers \cite{RT1, RT2}, vision-language-action (VLA) models \cite{VIMA, OpenVLA, Octo, RDT1b, pi0, Gr2}, and video-language planning models \cite{interactive-real-world-sim, vlp}. While these methods have shown promise for robot arms with grippers, the lack of large-scale datasets for humanoid robots -- particularly those with dexterous hands and active cameras -- remains a significant barrier to developing FMs for humanoid robots.

\textbf{Vision-Language Models (VLMs)} build upon the remarkable success of Large Language Models \cite{gpt4} to develop capabilities in multi-modal understanding and reasoning. Recent advancements include the development of text-image VLMs \cite{alayrac2022flamingo, chen2023minigpt,li2023blip, bai2023qwen, liu2024visual} and text-video VLMs \cite{zhang2023video, shu2023audio, maaz2023video, jin2024video}. In this work, we utilize the open-source VideoLLaMA2 \cite{videollama2} to train the Connector module within the humanoid agent, enhancing efficiency and grounding decision-making for embodied tasks.

\section{Conclusion and Limitations}

In this work, we introduced Being-0, a hierarchical agent framework for humanoid robots, designed to control a humanoid equipped with dexterous hands and active vision to solve long-horizon embodied tasks. The novel VLM-based Connector module effectively bridges the gap between the high-level Foundation Model and low-level skills, significantly enhancing the performance and efficiency of the humanoid agent. Extensive real-world experiments demonstrate Being-0's strong capabilities in navigation, manipulation, and long-horizon task-solving. The results highlight the effectiveness of the proposed Connector, the adjustment method for coordinating navigation and manipulation, and the use of active vision.

Despite these advancements, the current system does not incorporate complex locomotion skills such as crouching, sitting, or jumping. These skills could extend the humanoid's functionality beyond flat-ground settings, enabling tasks like climbing stairs, working from seated positions, or manipulating objects at varying heights. Enhancing these capabilities will be an important direction for future work. Additionally, while the onboard system is efficient, Being-0 still relies on the slow Foundation Model for high-level decision-making. Future research could explore lightweight Foundation Models tailored for robotics applications to further improve the system's efficiency.




\section*{Impact Statement}
This work explores advancements in humanoid robotic agents, which come with specific safety concerns. The use of Foundation Models and skill libraries introduces the potential risks of predicting incorrect skills or executing actions in out-of-distribution scenarios. For full-sized humanoid robots, such errors could lead to damage to surroundings or harm to people. At present, these systems should be tested only in controlled, experimental environments to ensure safety. Future work should prioritize robust error handling, fail-safes, and ethical guidelines to mitigate these risks and enable safer deployment of humanoid agents.




\bibliography{example_paper}
\bibliographystyle{icml2025}

\newpage
\appendix
\onecolumn
\section{Additional Results}

In this section, we provide detailed additional results from our experiments. Video recordings are included in the \textbf{supplementary material}.

\subsection{First-Person Video Records}

\begin{figure}[htbp]
\centering
\includegraphics[width=.95\linewidth, trim={0cm, 8cm, 2.5cm, 0cm}, clip]{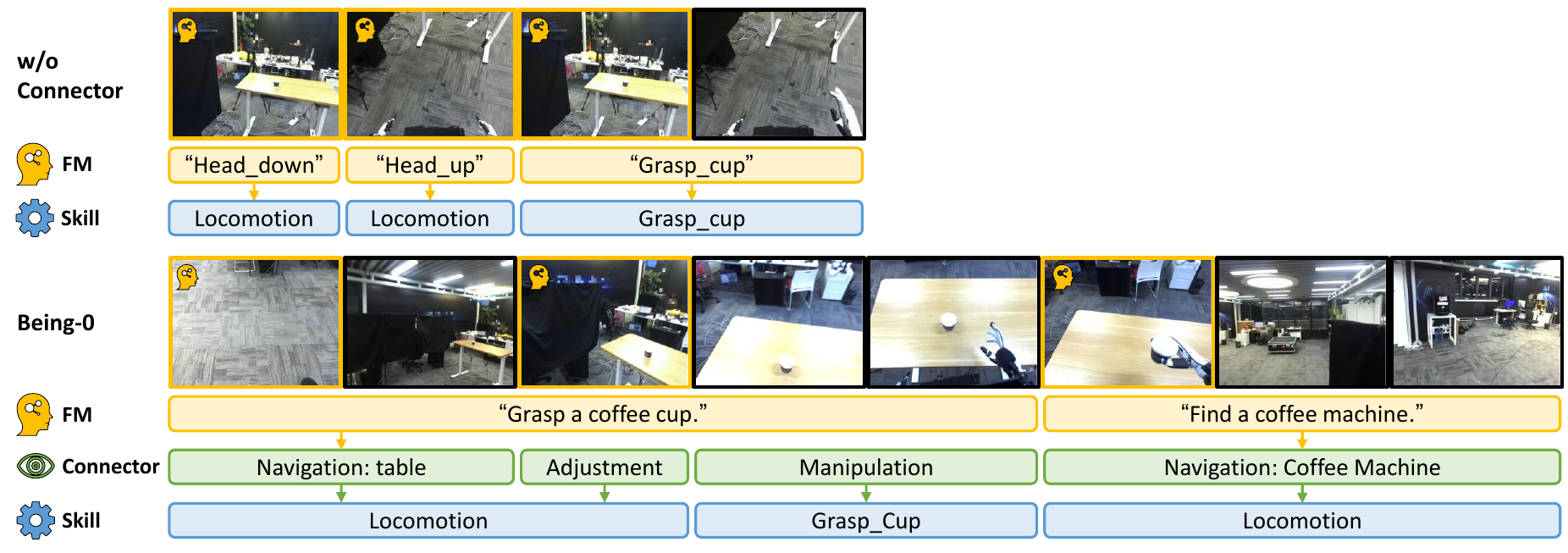}
\caption{A comparison of Being-0 w/o Connector and Being-0 in the long-horizon task ``Prepare-coffee." The first row shows recordings of Being-0 without the Connector, while the second row shows recordings of Being-0 with the Connector. Being-0 w/o Connector frequently queries the FM, which often fails to provide correct plans due to its limited embodied scene understanding. In contrast, Being-0 with the Connector completes the task, requiring only a few queries to the FM.}
\label{fig:app-wo-connector}
\end{figure}

\begin{figure}[htbp]
\centering
\includegraphics[width=.95\linewidth, trim={0cm, 7cm, 8cm, 0cm}, clip]{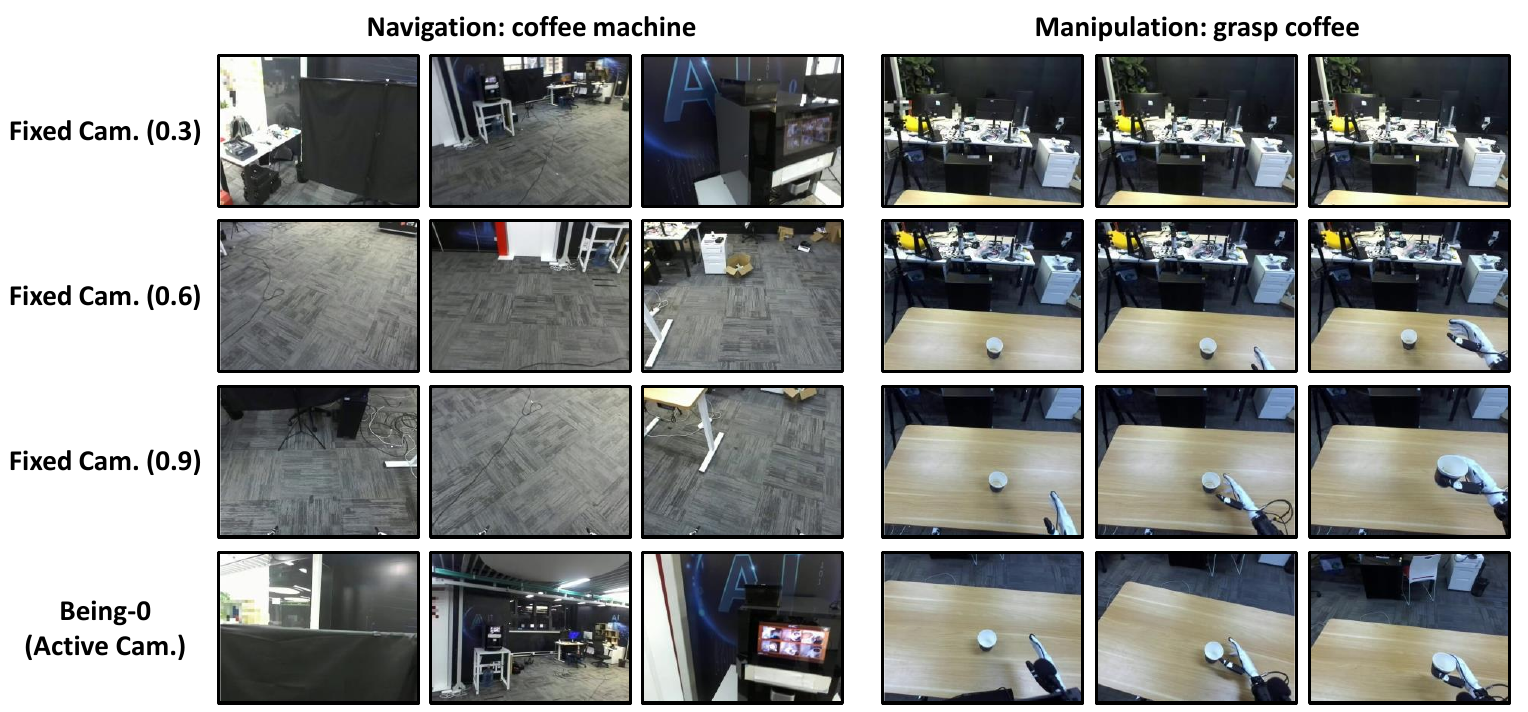}
\caption{Recordings from the ablation study on the active camera. Each row represents a different camera configuration, with the left three images depicting the navigation task and the right three images depicting the manipulation task. Only Being-0 with an active camera achieves robust performance in both navigation and manipulation.}
\label{fig:app-active-cam}
\end{figure}

\begin{figure}[htbp]
\centering
\includegraphics[width=.95\linewidth, trim={0cm, 9cm, 8cm, 0cm}, clip]{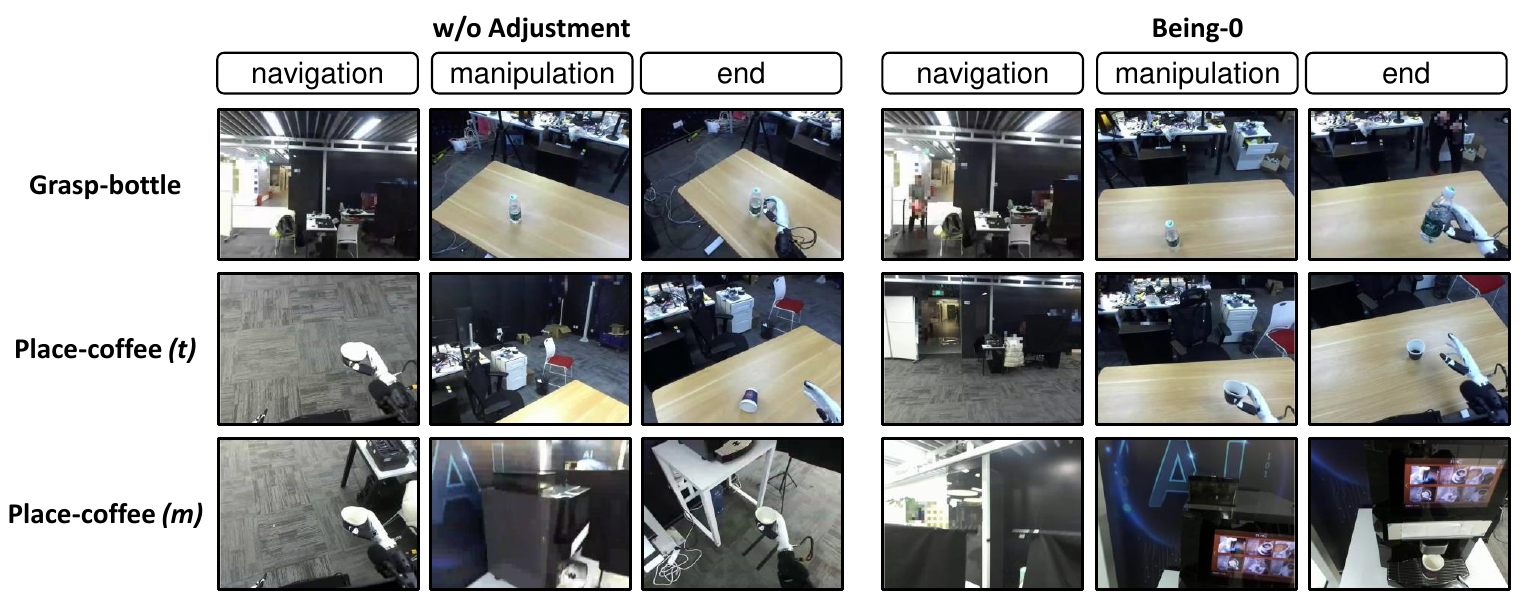}
\caption{A comparison of Being-0 with and without the adjustment method in two-stage tasks involving navigation and manipulation. Each row corresponds to a specific task, with the left three images showing results for Being-0 w/o Adjustment and the right three images showing results for Being-0. Without adjustment, the agent may terminate navigation in improper poses, leading to failed manipulations.}
\label{fig:app-wo-adjust}
\end{figure}

\begin{figure}[htbp]
\centering
\includegraphics[width=.95\linewidth, trim={0cm, 2cm, 11.5cm, 0cm}, clip]{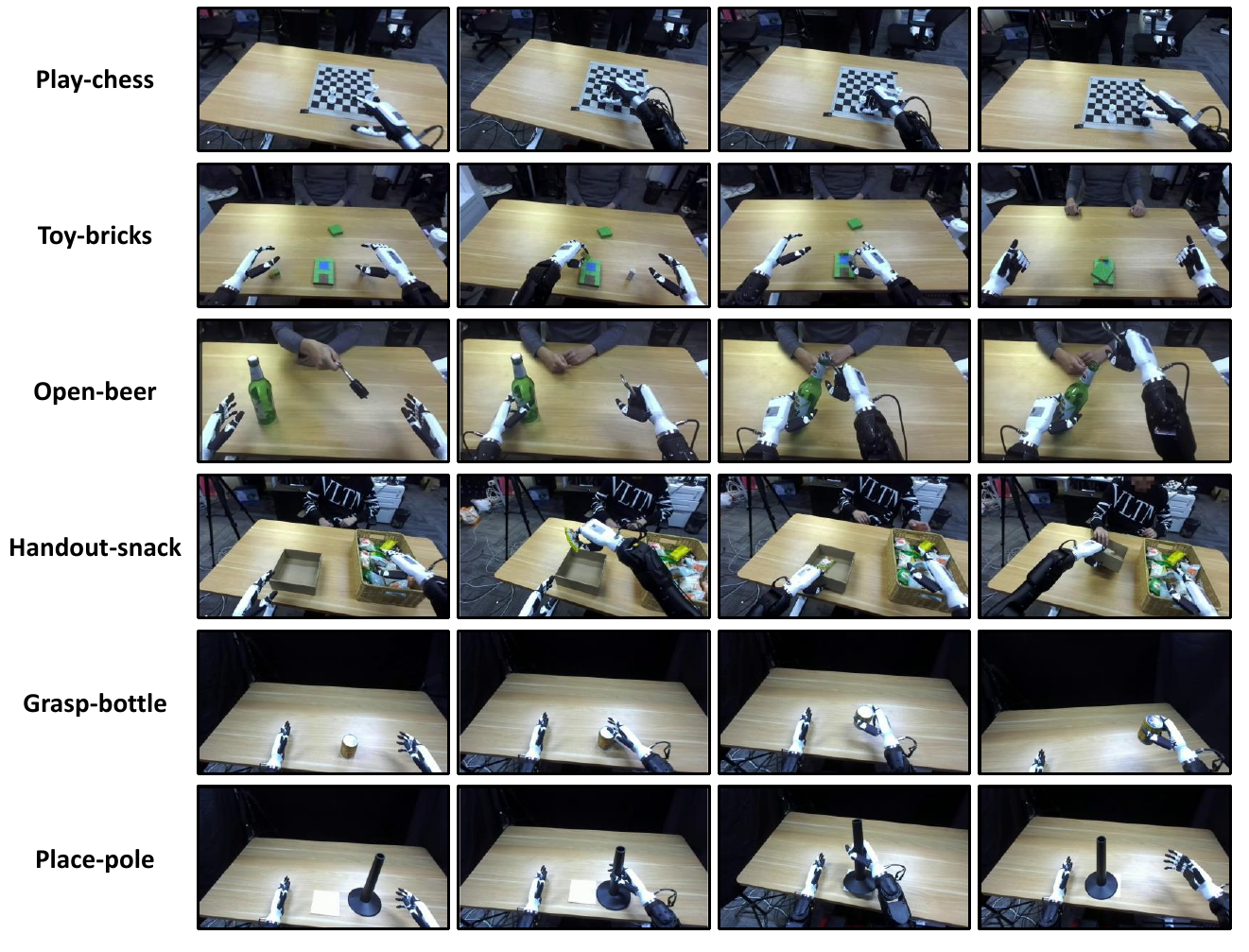}
\caption{First-person view recordings of the learned manipulation skills. Each row corresponds to a specific skill, with images from left to right depicting the progression of the manipulation process.}
\label{fig:app-manipulation}
\end{figure}

\newpage
\subsection{Foundation Model Planning Traces}

\begin{figure}[H]
\centering
\includegraphics[width=.95\linewidth, trim={0cm, 7cm, 0cm, 0cm}, clip]{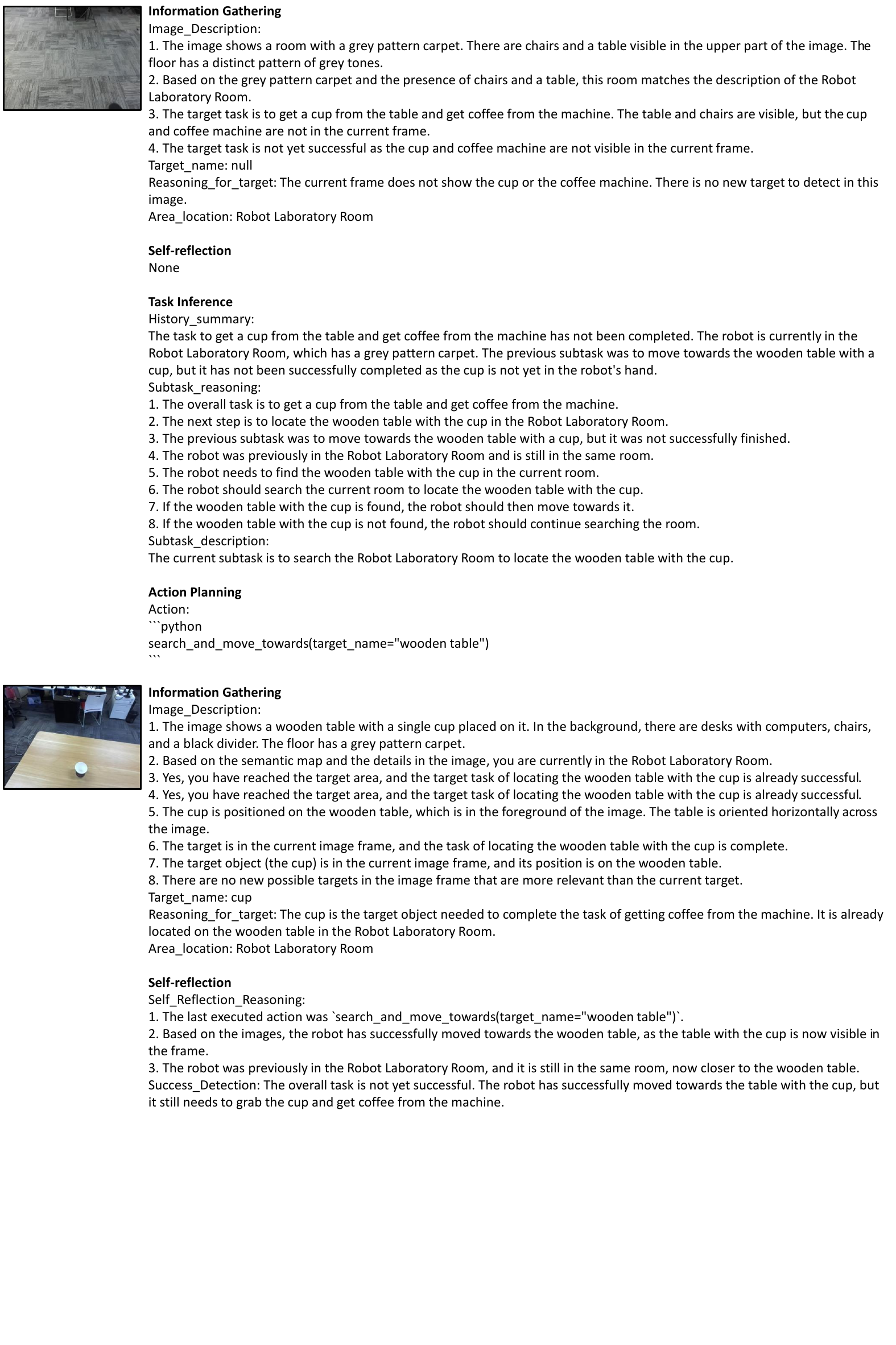}
\caption{Planning traces of the Foundation Model in \textbf{Being-0} for the task ``Prepare-coffee."}
\label{fig:app-trace-cup-1}
\end{figure}

\begin{figure}[htbp]
\centering
\includegraphics[width=.95\linewidth, trim={0cm, 7.5cm, 0cm, 0cm}, clip]{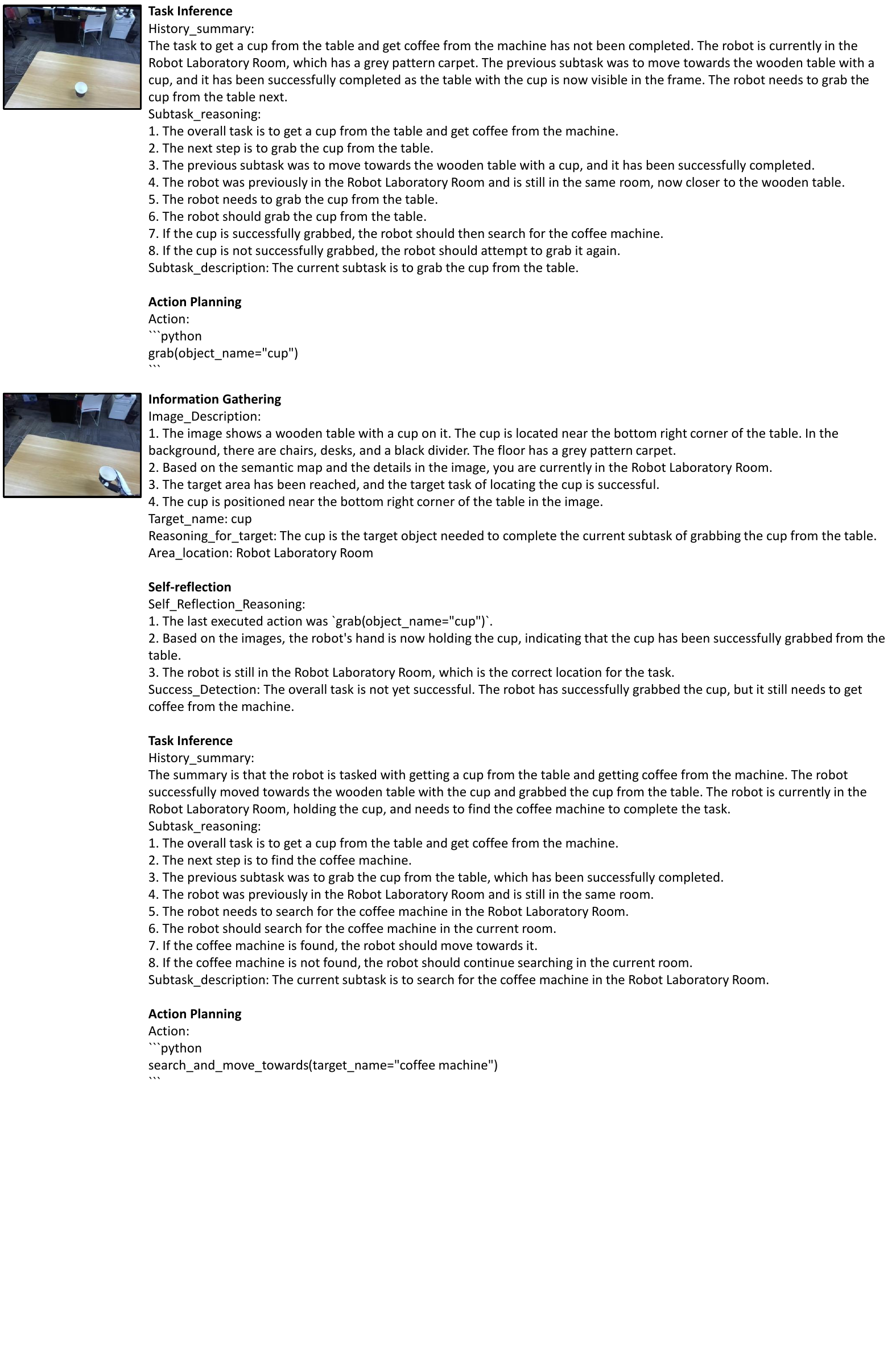}
\caption{(Continued) Planning traces of the Foundation Model in \textbf{Being-0} for the task ``Prepare-coffee."}
\label{fig:app-trace-cup-2}
\end{figure}

\begin{figure}[htbp]
\centering
\includegraphics[width=.9\linewidth, trim={0cm, 3cm, 0cm, 0cm}, clip]{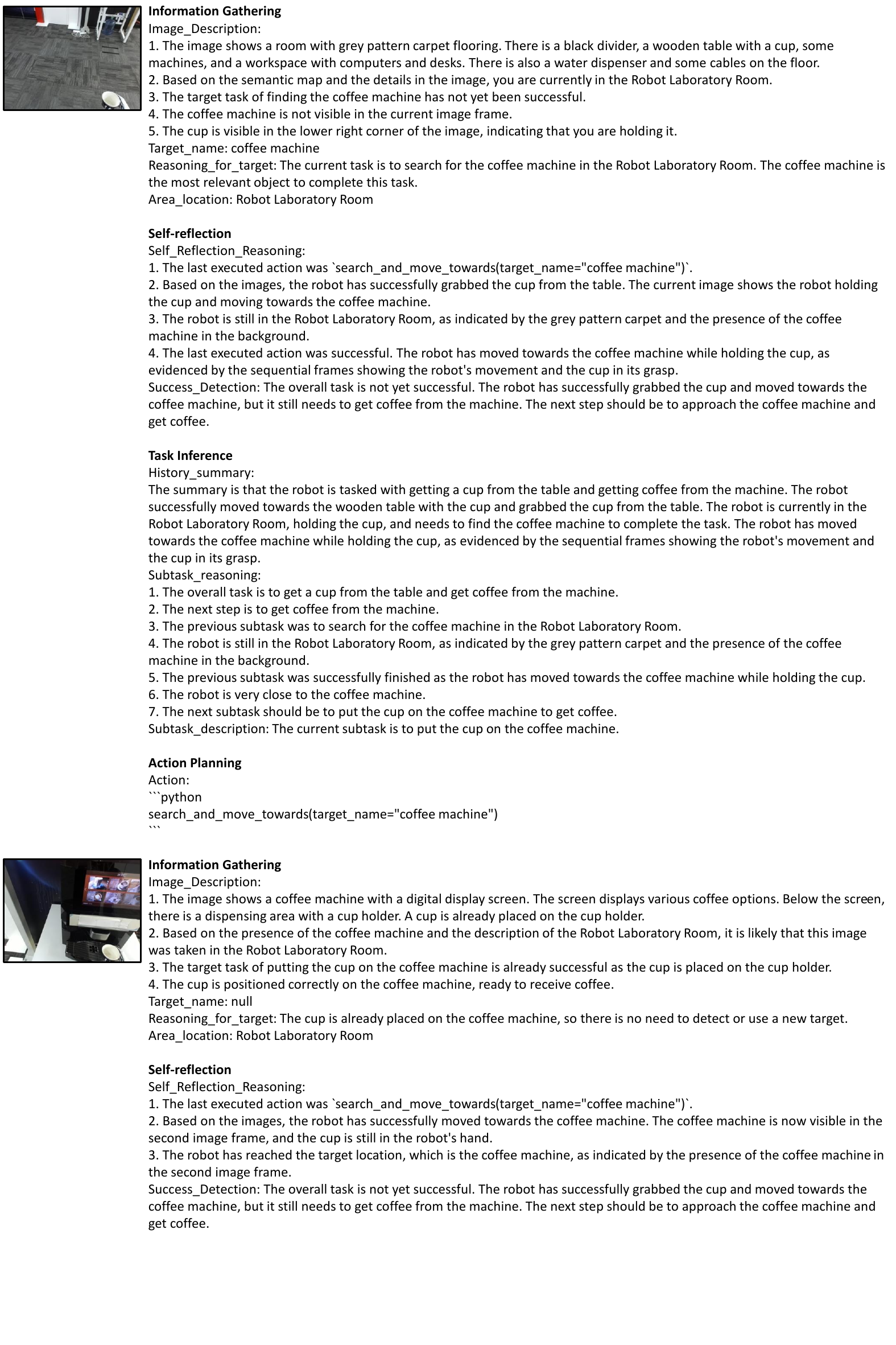}
\caption{(Continued) Planning traces of the Foundation Model in \textbf{Being-0} for the task ``Prepare-coffee."}
\label{fig:app-trace-cup-3}
\end{figure}

\begin{figure}[htbp]
\centering
\includegraphics[width=.9\linewidth, trim={0cm, 2.5cm, 0cm, 0cm}, clip]{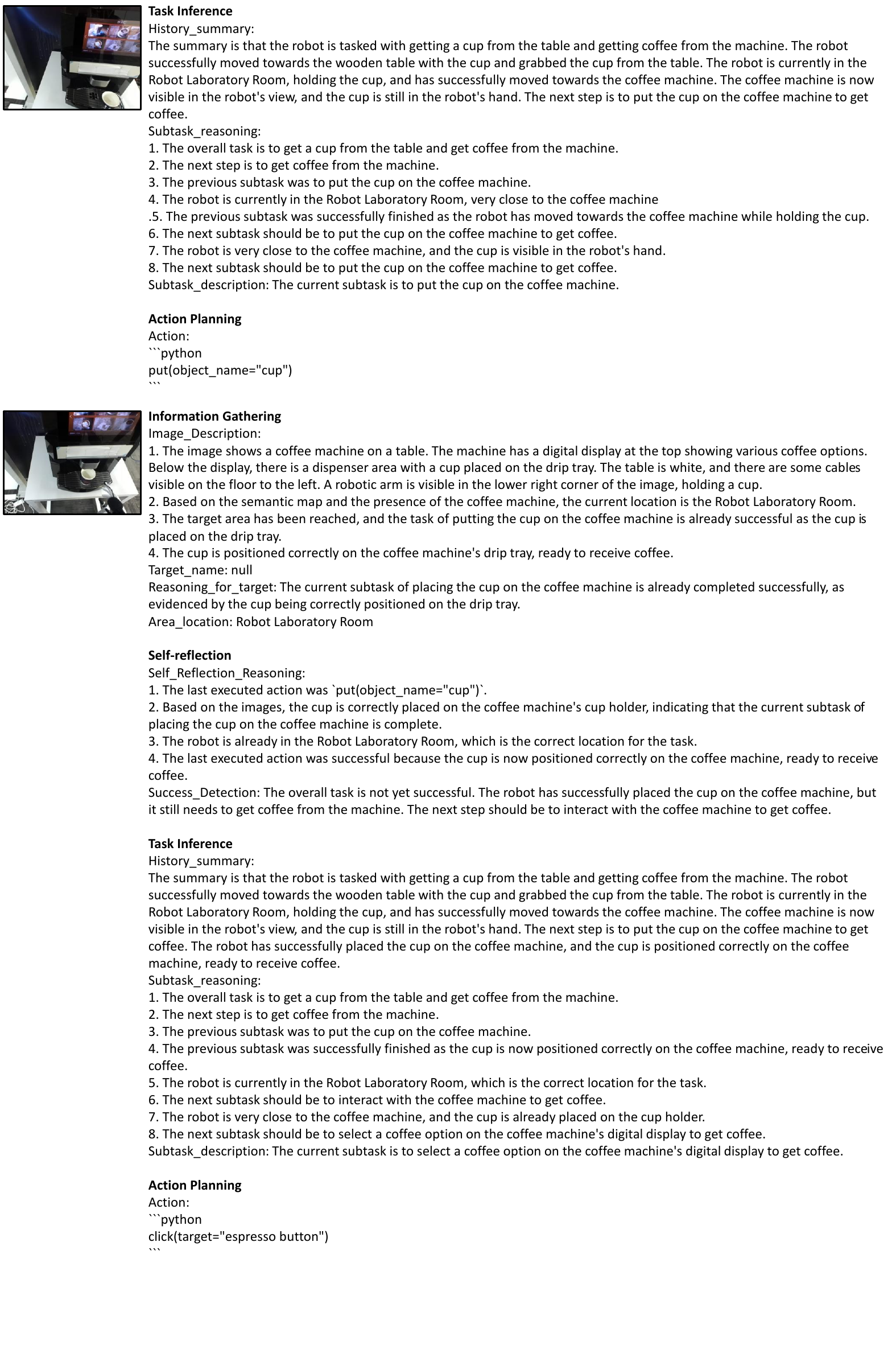}
\caption{(Continued) Planning traces of the Foundation Model in \textbf{Being-0} for the task ``Prepare-coffee."}
\label{fig:app-trace-cup-4}
\end{figure}



\begin{figure}[htbp]
\centering
\includegraphics[width=.95\linewidth, trim={0cm, 7cm, 0cm, 0cm}, clip]{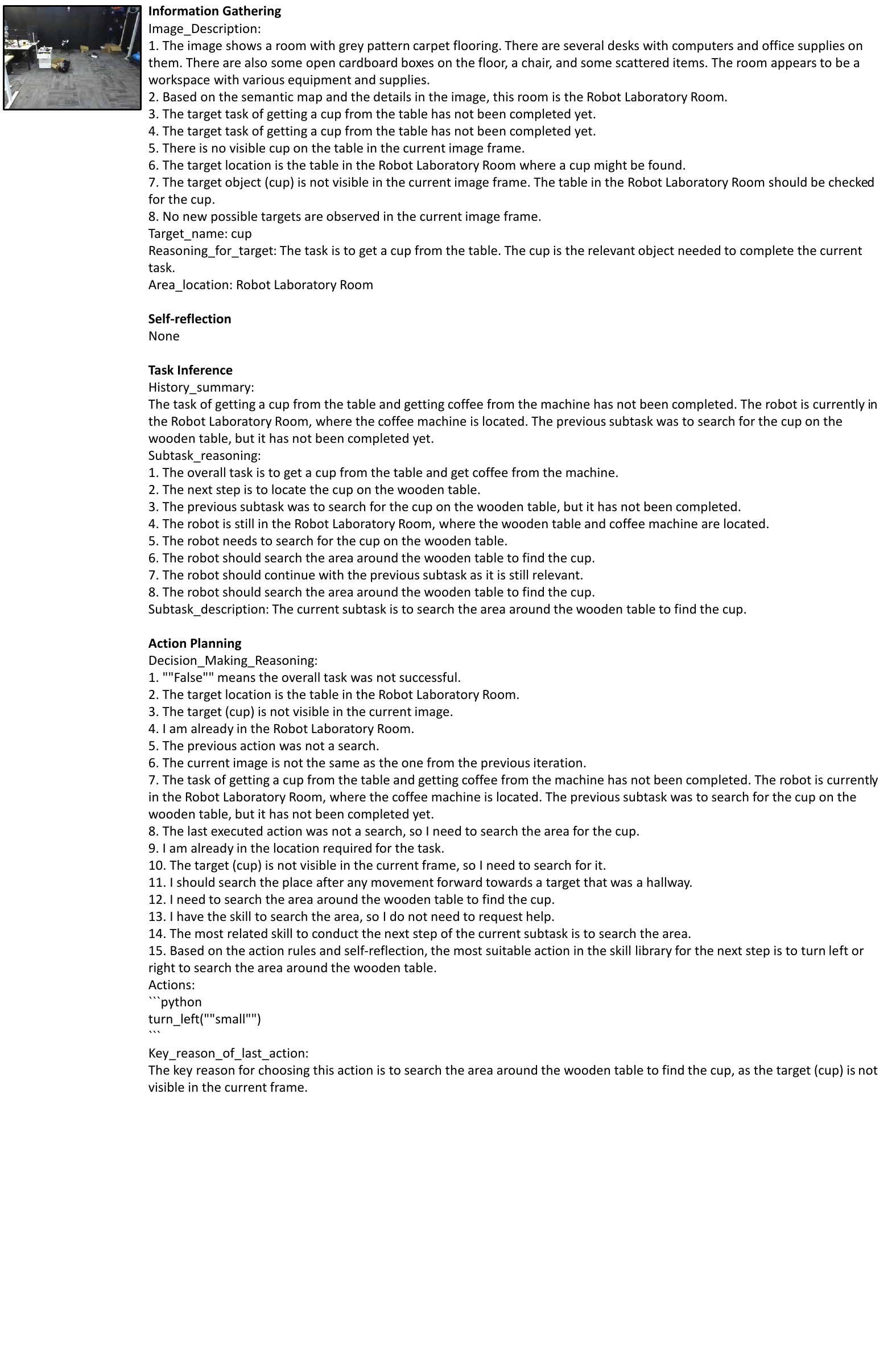}
\caption{Planning traces of the Foundation Model in \textbf{Being-0 w/o Connector} for the task ``Prepare-coffee."}
\label{fig:app-trace-woconnector-1}

\end{figure}
\begin{figure}[htbp]
\centering
\includegraphics[width=.9\linewidth, trim={0cm, 3.5cm, 0cm, 0cm}, clip]{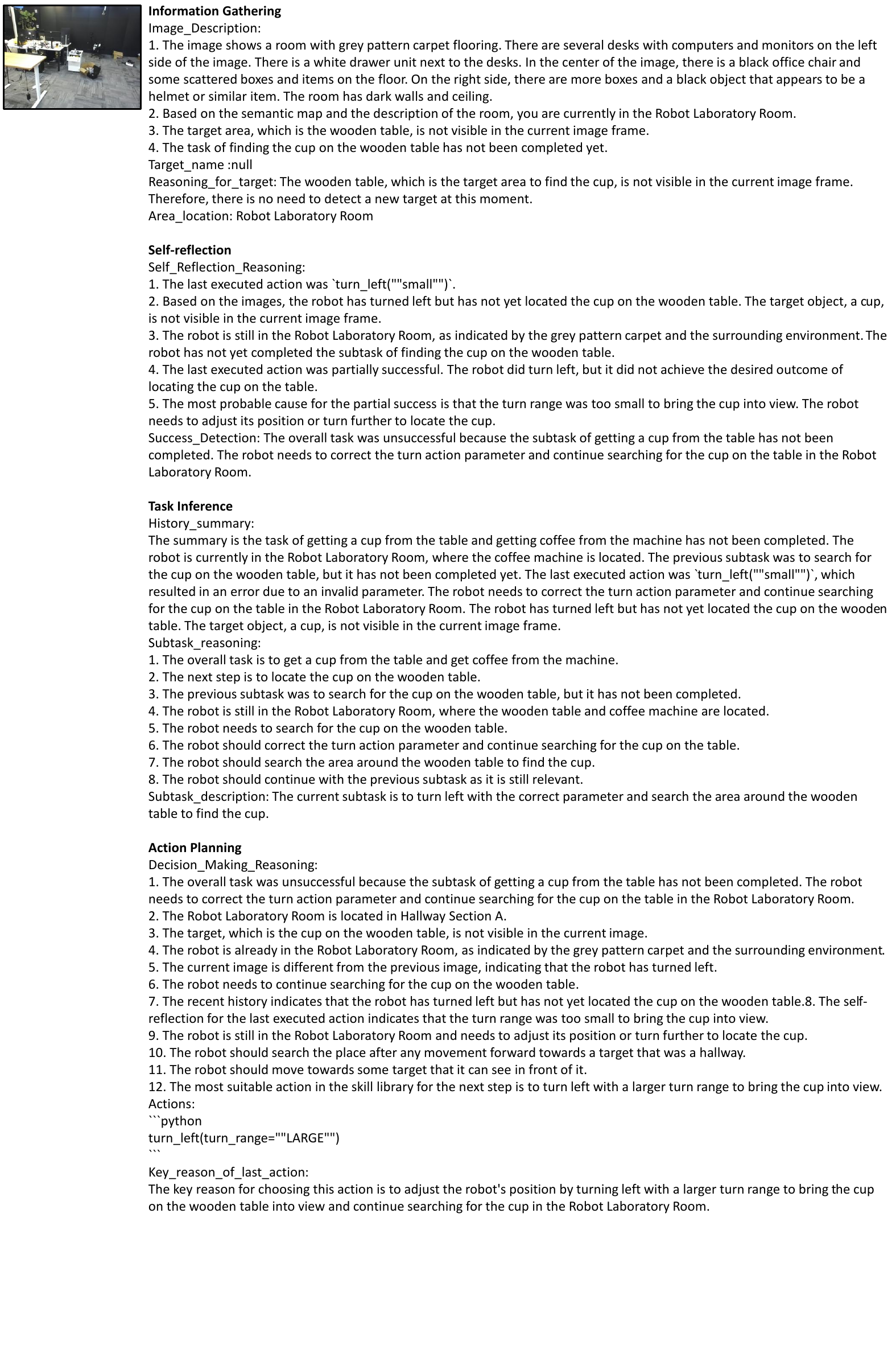}
\caption{(Continued) Planning traces of the Foundation Model in \textbf{Being-0 w/o Connector} for the task ``Prepare-coffee."}
\label{fig:app-trace-woconnector-2}
\end{figure}

\begin{figure}[htbp]
\centering
\includegraphics[width=.88\linewidth, trim={0cm, 1cm, 0cm, 0cm}, clip]{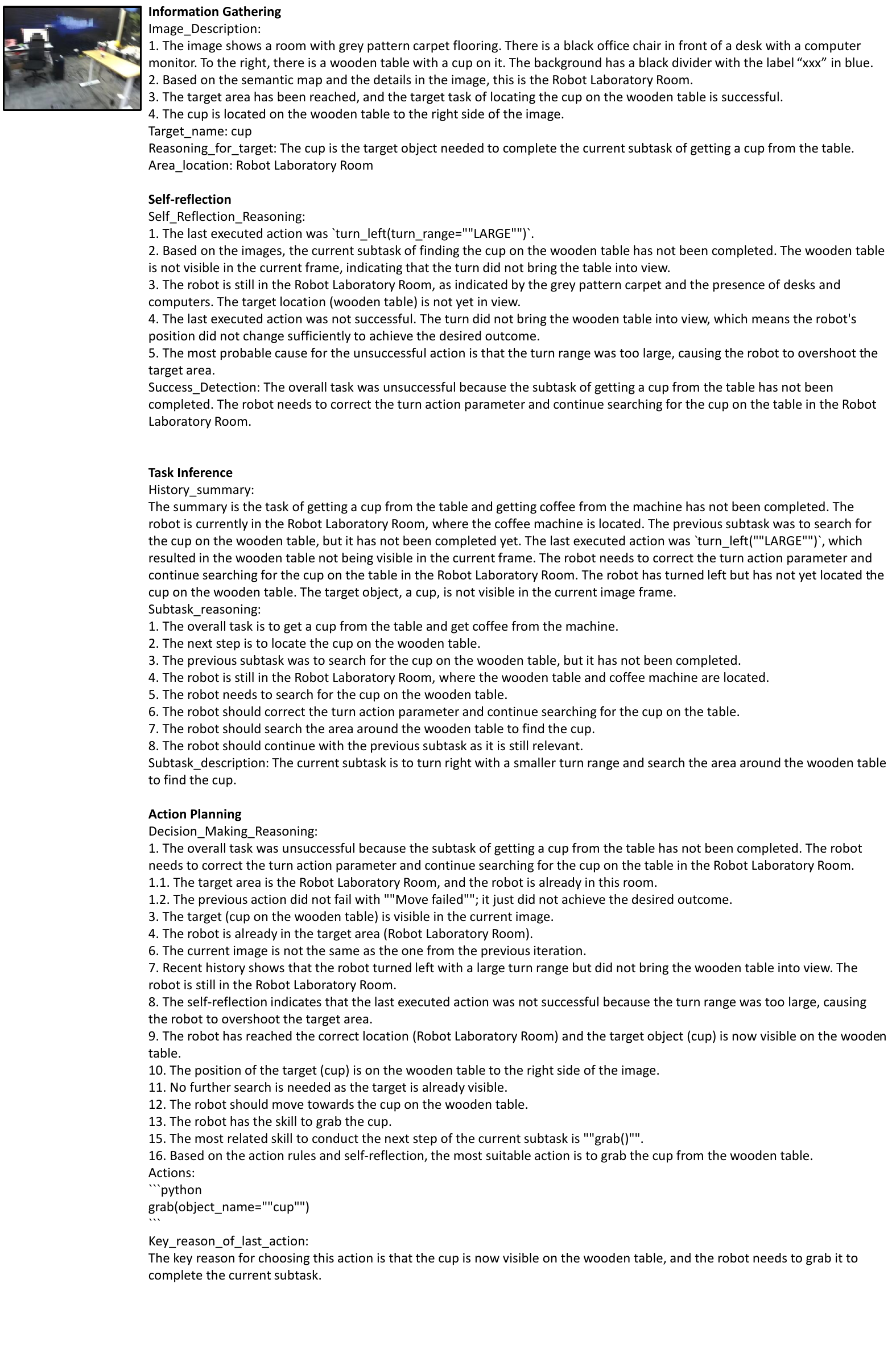}
\caption{(Continued) Planning traces of the Foundation Model in \textbf{Being-0 w/o Connector} for the task ``Prepare-coffee."}
\label{fig:app-trace-woconnector-3}
\end{figure}





\newpage
\subsection{Details of Experimental Setup and Results}
\label{app:exp-detail}


\begin{table}[htbp]
\renewcommand\arraystretch{1.2}
\caption{Detailed sub-processes required to complete each long-horizon task, along with the success rates of Being-0 and the baseline.}
\vspace{0.2cm}
\centering
\small
\begin{tabular}{llcc} 
\toprule
\textbf{Task} & \textbf{Sub-Process} & \textbf{\tabincell{c}{{w/o Connector}}}  &  \textbf{\tabincell{c}{{Being-0}}} \\ \midrule
\multirow{2}{*}{\textbf{Fetch-bottle}} & Navigate to table.  & 0 / 5 & 5 / 5 \\  
{}  & Grasp cup. & 0 / 5 & 4 / 5 \\ \midrule
\multirow{2}{*}{\textbf{Deliver-basket}} & Navigate to table.  & 0 / 5 & 5 / 5 \\  
{}  & Place basket. & 0 / 5 & 3 / 5 \\ \midrule
\multirow{4}{*}{\textbf{Prepare-coffee}} & Navigate to table.  & 0 / 5 & 5 / 5 \\  
{}  & Grasp cup. & 0 / 5 & 4 / 5 \\
{}  & Navigate to coffee machine. & 0 / 5 & 3 / 5 \\
{}  & Place cup. & 0 / 5 & 3 / 5 \\\midrule
\multirow{4}{*}{\textbf{Make-coffee}} & Place cup.  & 5 / 5 & 5 / 5 \\  
{}  & Select coffee. & 5 / 5 & 5 / 5  \\
{}  & Select confirmation. & 4 / 5 & 4 / 5  \\
{}  & Grasp cup. & 4 / 5 & 4 / 5  \\\midrule
\multirow{3}{*}{\textbf{Deliver-coffee}} & Grasp-cup.  & 5 / 5 & 5 / 5 \\  
{}  & Navigate to table. & 0 / 5 & 4 / 5 \\
{}  & Place cup. & 0 / 5 & 4 / 5 \\
\bottomrule 
\end{tabular}
\end{table}

\section{Implementation Details}

\subsection{Acquiring Manipulation Skills}
\label{app:manip-skills}

Table \ref{tab:manipulation_data} presents the number of successful trajectories collected for each skill via teleoperation. In the ACT policy for each skill, we utilize a ResNet-50 backbone pre-trained on ImageNet to process binocular images. To enhance robustness against visual perturbations, data augmentation techniques such as random cropping, rotation, and color jittering are applied. The entire model, including the pre-trained encoder, is updated during training. Table \ref{tab:manipulation_train} lists the hyperparameters used for training ACT.

\begin{table}[htbp]
\caption{Number of trajectories collected for each manipulation skill.}
\label{tab:manipulation_data}
\centering
\begin{tabular}{l r}
\toprule
\textbf{Skill} & \textbf{Num. Trajectories} \\
\midrule
Carry Basket & 25 \\
Handout Snack & 50 \\
Grasp Bottle & 150 \\
Grasp Cup & 200 \\
Open Beer & 50 \\
Place Basket & 25 \\
Place Cup & 200 \\
Place Pole & 50 \\
Play Chess & 70 \\
Play Toy Bricks & 50 \\
\bottomrule
\end{tabular}
\end{table}

\begin{table}[htbp]
\caption{Hyperparameters used for training the ACT policy.}
\label{tab:manipulation_train}
\centering
\begin{tabular}{l r}
\toprule
\textbf{Hyperparameter} & \textbf{Value} \\
\midrule
Training steps & 500,000 \\
Batch size & 90 \\
Learning rate & 1e-5 \\
Gradient clip norm & 10 \\
Chunk size (train) & 30 \\
Chunk size (test) & 10 \\
\bottomrule
\end{tabular}
\end{table}

\subsection{The Embodied Connector}
\label{app:connector}

\subsubsection{Dataset Statistics}
Our dataset consists of two major types of tasks: \textbf{Visual Understanding (VLU) tasks} and \textbf{Action Planning (AP) tasks}. The VLU tasks include bounding box detection, yes/no questions, and image description tasks. 
We collected a total of 3,177 images, with 2,483 images dedicated to visual understanding tasks and 694 images for action planning tasks.
Specifically, the image description tasks were initially labeled by GPT-4o and then refined by human annotators to ensure accuracy. The examples of the visual understanding labeling tasks are presented in Table~\ref{tab:task_samples}.
Table~\ref{tab:task_counts} summarizes the number of samples available for each type of task. Table~\ref{tab:task_counts_bounding_box} and \ref{tab:task_counts_yesno} shows the data statistics of the bounding box tasks and the yes/no tasks, respectively.

{To enhance visual grounding capabilities, we also include a general visual grounding dataset. We filtered 300K data samples from three open-source visual grounding datasets \citep{krishna2016visual,tian2024chatterbox,shao2024visual}.}
These data are fed into the model prior to training on our VLU and AP datasets, which will be further mentioned in Appendix~\ref{app:training_details}.









\begin{table}[htbp]
\caption{Examples of the training data for training the vision-language model to acquire the visual understanding ability.}
\label{tab:task_samples}
\centering
\resizebox{0.99\linewidth}{!}{
\begin{tabular}{|l|l|}
\toprule
\textbf{Task Name} & \textbf{Samples} \\
\midrule
Bounding Boxes & \textbf{Text Prompt}: Detect the bounding box of target object: Reception desk. \\
~ & \textbf{Answer}: [0.471, 0.082, 0.765, 0.951] \\
\midrule
Identify items & \textbf{Text Prompt}: Is there an exit in the image? Do not generate anything else except 'Yes' or 'No'. \\
~ & \textbf{Answer}: No. \\
\midrule
Describe image & \textbf{Text Prompt}: Describe the current image. List all the items you could see. \\
 & \textbf{Answer}: The image shows a corridor or office area with the following items visible: \\
 &  1. A person wearing a light-colored outfit. \\
 &  2. A green-framed glass door. \\
 &  3. A yellow sofa with cushions. \\
 &  4. A wall-mounted tablet or screen. \\
\midrule
Describe the ground and identify the transition & \textbf{Text Prompt}: Describe the ground and identify the transition between different grounds if there are any. \\
& \textbf{Answer}: grey pattern carpet \\
\bottomrule
\end{tabular}
}

\end{table}

\begin{table}[htbp]
\caption{Task Categories and Sample Numbers}
\label{tab:task_counts}
\centering
\resizebox{0.75\linewidth}{!}{
\begin{tabular}{ll|l}
\toprule
\textbf{Task categories} & \textbf{Tasks} & \textbf{Sample number} \\
\midrule
\multirow{4}{*}{Visual Understanding} 
 & Bounding boxes &  14,784 \\
 & Identify items & 20,536 \\
 & Describe image & 1,530 \\
 & Describe the ground and identify the transition & 1,530 \\
\midrule
Action Planning & - & 771 \\
\bottomrule
\end{tabular}
}

\end{table}

\begin{table}[htbp]
\caption{Category data overview for the bounding box task.}
\label{tab:task_counts_bounding_box}
\centering
\resizebox{0.35\linewidth}{!}{
\begin{tabular}{l r}
\toprule
\textbf{Category} & \textbf{Value} \\
\midrule
Kitchen Area & 792 \\
Hallway & 3,834 \\
Robot Laboratory Room & 896 \\
Reception Area & 247 \\
Coffee machine & 1,323 \\
Workspace & 118 \\
Meeting Room & 1,590 \\
Wooden Table & 414 \\
Closed door & 727 \\
Workspace Area & 697 \\
Reception Desk & 1,026 \\
Door label & 513 \\
Doorway & 706 \\
Reception & 114 \\
Digital screen & 428 \\
Cup area & 338 \\
Espresso coffee button & 496 \\
Confirm button & 286 \\
Cancel button & 239 \\
\midrule
\textbf{Total} & \textbf{14,784} \\
\textbf{Number of Categories} & \textbf{19} \\
\bottomrule
\end{tabular}
}
\end{table}
\begin{table}[htbp]
\caption{Category data overview for the yes/no task.}
\label{tab:task_counts_yesno}
\centering
\resizebox{0.35\linewidth}{!}{
\begin{tabular}{l r}
\toprule
\textbf{Category} & \textbf{Value} \\
\midrule
Coffee machine & 1,530 \\
Reception desk & 1,530 \\
Closed door & 1,530 \\
Door label & 1,530 \\
Water fountain & 1,530 \\
Glass door & 1,530 \\
Hallway & 1,530 \\
Reception area & 1,530 \\
Exit & 1,530 \\
Workspace & 1,530 \\
Passage & 1,530 \\
Doorway & 1,530 \\
Digital screen & 428 \\
Espresso coffee button & 428 \\
Confirm button & 446 \\
Preparing screen & 428 \\
Coffee ready screen & 428 \\
Cancel button & 18 \\
\midrule
\textbf{Total} & \textbf{20,536} \\
\textbf{Number of Categories} & \textbf{18} \\
\bottomrule
\end{tabular}
}
\end{table}
\subsubsection{Training Details}
\label{app:training_details}




We fine-tuned our vision-language model using the VideoLLaMA2 framework with a multi-node distributed training strategy. The training was conducted with a global batch size of 128 and a local batch size of 2 per device, with gradient accumulation steps dynamically computed based on the number of nodes and processing units per node. The model was trained for three epochs using a learning rate of $2 \times 10^{-5}$, a cosine learning rate scheduler, and a warmup ratio of 0.03. We employed AdamW as the optimizer with zero weight decay.
To enhance computational efficiency, we enabled mixed precision training with bfloat16 (BF16) and TensorFloat32 (TF32). Gradient checkpointing was applied to reduce memory consumption, and the maximum sequence length was set to 4096 tokens.
The vision encoder was based on SigLIP\footnote{\url{https://huggingface.co/google/siglip-so400m-patch14-384}}~\citep{zhai2023sigmoid}, while the projection module was implemented using a multi-layer perceptron (MLP). We grouped multimodal samples by modality length, selected vision features from the second-to-last layer, and applied image padding to maintain aspect ratios. Each sample contained 16 frames.

{The overall training process is divided into two stages. First, we finetune the model with the filtered 300K general visual grounding dataset based on the checkpoint provided by VideoOrion$+$\citep{feng2024videoorion}, which shares the same architecture with VideoLLaMA2 but offers better object-centric understanding capabilities. Considering the training efficiency, we modify VideoOrion by removing the object-centric branch.  Then we finetune the resulting model with our collected dataset including VLU tasks and AP tasks together.}

This training setup ensures efficient vision-language modeling, leveraging optimized data handling, memory-efficient techniques, and distributed computation for improved performance.


\subsubsection{Usage of the Connector}

\textbf{Visual Understanding.}
For visual understanding , the trained model predicts the bounding boxes of target objects in an image by generating the coordinates of the box, or it outputs “None” if no target object is present. The visual understanding capability of the VLM provides the robot with concrete information about its environment, enabling effective navigation and laying the foundation for informed skill planning decisions.

\textbf{Skill Planning.}
For skill planning, given an overall task and a subtask, the model predicts the appropriate skill code for the robot to execute from either the modular skill library or the composite navigation skills. Skill planning facilitates the acquisition of spatial and embodied knowledge by enabling the model to make decisions based on the presence, relative distances and positioning of objects within its environment. Through training, the model learns to assess these spatial relationships and select the appropriate skill such as navigation and manipulation according to the proximity and orientation of the target objects. This embodied understanding allows the robot to adapt its actions in real-time, ensuring the translation of high-level task instructions into precise, contextually relevant actionable skills.

Below, we present details of the composite locomotion skills for navigation.

\textbf{Move towards.} We define the move towards skill as a skill that help the robot navigate to an target object in its view. The bounding box generated by the VLM is leveraged in this skill to determine the angle and the existence of the target object. The pseudo code for this skill is shown in Algorithm~\ref{alg:move_towards}.

\begin{algorithm}[htbp]
\caption{Move Towards Target}
\label{alg:move_towards}
\begin{algorithmic}[1]
\STATE \textbf{Input:} \texttt{target\_name}
\STATE \textbf{Output:} Status of movement towards the target
\STATE Initialize: \texttt{max\_iterations}, \texttt{angle\_threshold}, \texttt{max\_iterations}
\FOR{each iteration from 1 to \texttt{max\_iterations}}
    \STATE Get camera image and depth data
    \STATE Detect the target object in the image using VLM
    \IF{No target detected}
        \STATE Stop moving
        \STATE Break the loop
    \ENDIF
    \IF{Target is within threshold distance}
        \STATE Stop moving
        \STATE Break the loop
    \ENDIF
    \IF{Obstacles detected}
        \STATE Avoid obstacle using sidestep
    \ELSE
        \STATE Move forward or turn depending on angle
    \ENDIF
\ENDFOR
\STATE \textbf{Return:} \texttt{status: True/False}
\end{algorithmic}
\end{algorithm}

\textbf{Search for.} We define the search for skill as constantly turning to one direction until the target object is found in the view. The bounding box generated by the VLM is leveraged in this skill to determine the existence of the target object. The pseudo code for this skill is shown in Algorithm~\ref{alg:search_for}.

\begin{algorithm}[htbp]
\caption{Search for Target}
\label{alg:search_for}
\begin{algorithmic}[1]
\STATE \textbf{Input:} \texttt{target\_name}
\STATE \textbf{Output:} Status of target search
\STATE Initialize: \texttt{max\_iterations}, \texttt{direction}, \texttt{head\_angle}, \texttt{tilt\_angle}
\FOR{each iteration from 1 to \texttt{max\_iterations}}
    \STATE Get camera image
    \STATE Check if target is detected in the image
    \IF{Target detected}
        \STATE Stop moving
        \STATE Break the loop
    \ENDIF
    \IF{Direction is "right"}
        \STATE Turn right
    \ELSE
        \STATE Turn left
    \ENDIF
\ENDFOR
\STATE \textbf{Return:} \texttt{status: True/False}
\end{algorithmic}
\end{algorithm}

\textbf{Adjustment.} To perform adjustment during navigation, we modify the move-forward skill so that the robot first adjusts its head to look aside to the direction of the item, then decides whether to turn or move forward based on the adjusted view. The direction to look aside is predicted by the VLM. This approach allows the robot to gradually approach the target object in an arc-shaped path, ultimately reaching the optimal position.  The pseudo code for this skill is shown in Algorithm~\ref{alg:final_adjust}.


\begin{algorithm}[htbp]
\caption{Adjustment}
\label{alg:final_adjust}
\begin{algorithmic}[1]
\STATE \textbf{Input:} \texttt{target\_name}, \texttt{direction}
\STATE \textbf{Output:} Status of final approach
\STATE Initialize: \texttt{head\_angle}, \texttt{tilt\_angle}, \texttt{max\_iterations}
\FOR{each iteration from 1 to \texttt{max\_iterations}}
    \STATE Set head position and tilt
    \STATE Get camera image and detect target direction
    \IF{Target detected and within threshold distance}
        \STATE Stop moving, adjust to face target
        \STATE Break the loop
    \ENDIF
    \STATE If target angle is small/large, adjust direction (left/right)
    \STATE If target angle is 0, check for obstacles and avoid if necessary
    \STATE Move towards the target if no obstacles detected
\ENDFOR
\STATE \textbf{Return:} \texttt{status: True/False}
\end{algorithmic}
\end{algorithm}

\newpage
\clearpage
\subsection{Prompt Design}
\label{app:prompt}

We present prompts designed to enable the foundation model to perform various levels of agent abilities. Specifically, we include prompts for information gathering, self-reflection, and subtask inference in Tables~\ref{tab:information_gathering}, \ref{tab:self-reflection}, and \ref{tab:subtask_reasoning}, respectively. Additionally, we provide the action planning prompt for scenarios where the embodied Connector is not used in Table~\ref{tab:action_planning}. For cases involving the embodied Connector, we utilize a more concise version of the prompt for action planning, which is shown in Table~\ref{tab:action_planning_short}.

\begin{table}[ht]
\caption{The prompt we used for information gathering process.}
\centering
\resizebox{0.99\linewidth}{!}{
\begin{tabular}{| p{20cm} |}
\toprule
\multicolumn{1}{|c|}{\textbf{Information Gathering}} \\
\midrule
You are a helpful AI assistant integrated with a humanoid robot body equipped to handle diverse tasks in the real world. Your advanced capabilities enable you to process textual and visual information, including computer application screenshots, and to control the robot body. \\
\midrule
\textbf{\texttt{<$image\_introduction$>}} \\
\midrule
\textbf{Overall task:} \\
\textbf{\texttt{<$task\_description$>}} \\
\midrule
\textbf{Subtask description:} \\
\textbf{\texttt{<$subtask\_description$>}} \\
\midrule
\textbf{Semantic map:} \\
\textbf{\texttt{<$semantic\_map$>}} \\
\midrule
\textbf{Current Location:} \\
\textbf{\texttt{<$robot\_location$>}} \\
\midrule
\textbf{Holding Cup Status:} \\
\textbf{\texttt{<$robot\_holding\_cup\_status$>}} \\
\midrule
\textbf{Image Description:} \\
1. Using the latest image, please describe it in detail. Pay attention to the details in the image, if any, especially critical objects or icons. \\
2. Identify in which area of the map you are currently in, based on the semantic map provided above and the past action history. \\
3. Pay attention if you have reached a target area and if the target task is already successful. \\
4. Pay attention if you have reached a target area and if the target task is already successful. \\
5. Keep in mind the target object or area and describe its position and orientation in the image, if any. \\
6. If you are trying to navigate to a location, use its place name from the semantic map as a target, if possible. \\
7. If the target is not in the current image frame, but it has been found previously, use the recent actions and the previous frames to reason about its position, location, and orientation. \\
8. In the latest image, if you observe new possible targets, compare the new targets with the current target and decide which one is more likely to be the correct target. \\
\midrule
\textbf{Target Name:} \\
Assume you can use a detection model to detect the most relevant object, image area, or UI item to complete the current task, if any is needed. What target should be detected to complete the task based on the latest image and the current task? You should obey the following rules: \\
1. Identify an item or area that is relevant to the current or intermediate target of the task. \\
2. For a target object, consider its possible forms and list as many as possible. \\
3. If there is a new possible target object, compare it with the current one and choose the one that is the most promising. \\
4. If there is no need to detect a target, only output ``null''. \\
\midrule
\textbf{Reasoning for Target:} \\
Why was this target chosen, or why is there no need to detect or use a new target? Why is this target more promising than other possible targets? \\
You should only respond in the format described below and not output comments or other information. DO NOT change the title of each item. \\
\textbf{Image Description:} \\
1. ... \\
2. ... \\
3. ... \\
4. ... \\
\textbf{Target Name:} \\
name \\
\textbf{Reasoning for Target:} \\
\texttt{...} \\
\textbf{Area Location:} \\
area name \\
\bottomrule
\end{tabular}
}

\label{tab:information_gathering}
\end{table}

\begin{table}[ht]
\caption{The prompt for summarizing task progress and proposing a new subtask.}
\centering
\resizebox{0.99\linewidth}{!}{
\begin{tabular}{| p{27cm} |}
\toprule
\multicolumn{1}{|c|}{\textbf{Task Reflection and Subtask Proposal}} \\
\midrule
\textbf{Overall task description:} \\
\texttt{<$task\_description$>} \\
\midrule
\textbf{Previous proposed subtask for the task:} \\
\texttt{<$subtask\_description$>} \\
\midrule
\textbf{Previous reasoning for proposing the subtask:} \\
\texttt{<$subtask\_reasoning$>} \\
\midrule
\textbf{\texttt{<$image\_introduction$>}} \\
\midrule
\textbf{Description of current image frame:} \\
\texttt{<$image\_description$>} \\
\midrule
\textbf{Last executed action:} \\
\texttt{<$previous\_action$>} \\
\midrule
\textbf{Error report for the last executed action:} \\
\texttt{<$executing\_action\_error$>} \\
\midrule
\textbf{Key decision-making reasoning for the last executed action:} \\
\texttt{<$previous\_reasoning$>} \\
\midrule
\textbf{Self-reflection for the last executed action:} \\
\texttt{<$self\_reflection\_reasoning$>} \\
\midrule
\textbf{Success Detection for the overall task:} \\
\texttt{<$success\_detection$>} \\
\midrule
\textbf{The following is the summary of history that happened before the last screenshot:} \\
\texttt{<$previous\_summarization$>} \\
\midrule
\textbf{Semantic map:} \\
\texttt{<$semantic\_map$>} \\
\midrule
\textbf{Current Location:} \\
\texttt{<$robot\_location$>} \\
\midrule
\textbf{Holding Cup Status:} \\
\texttt{<$robot\_holding\_cup\_status$>} \\
\midrule
\textbf{History summary:} \\
Summarize what happened previously, especially the last step according to the decision-making reasoning and self-reflection reasoning for the last executed action. The summary needs to be precise, concrete, highly related to the task, and follow the rules below. \\
    1. Determine if the task has been completed successfully. If it is successful, ignore question 2 to 5. \\ 
    2. Summarize the tasks from the history and the current task. What is the current progress of the task? For example, to open a file, you first need to select the file, then open it by clicking somewhere or using the keyboard. Subtasks may have other pre-requisites. \\
    3. Record the successful actions and organize them into events, step by step. \\
    4. What is the current area you are in? What is the target area? What is the next area going to be if you move forward? \\
    5. Which subtask has been completed? Which subtasks have not been completed? \\
    6. Do not forget the information and key events in the previous steps of the overall task. \\
\midrule
\textbf{Subtask reasoning:} \\
1. Based on the unfinished part of overall task and the current visual information, identify the way to complete the task without making any assumptions beyond the provided information. \\
2. Analyze the target task step by step to determine how to complete it. \\
3. What was the previous subtask? Was the previous subtask successfully finished according to self-reflection? Is it improper for the current situation? If finished or improper, please select a new subtask, otherwise you must reuse the last subtask. \\
4. What was the previous location you were in? Have you reached a new place based on the current observation? Pay attention if you have already reached the target location of the previous subtask or task. \\
5. If you are already in a new location, PLEASE propose a new subtask and skip question 14. \\
6. If the target of the action is not visible in the current image, DO NOT try to move towards it. Instead, if you are in the target location, the new subtask should be to search around to find it. \\
7. You should search the place for the target after any movement towards a previous target. \\
8. If the search does not find the target, propose a new task to move towards a new target in the current image, and then search again for the previous target. \\
9. If the next area in front of you is not the target area, DO NOT move towards it. Instead, if you are in the target location, propose a new subtask to search to find it. \\
10. If you want to propose a new subtask, give reasons why it is more feasible for the current situation. Please strictly follow the description and requirements in the current task. \\
11. The proposed subtask needs to be precise and concrete within one sentence. \\
12. If a given task or subtask is already very simple, like "wave your hand", no need to decompose it, the next subtask to perform is just the simple task. \\
\midrule
You should only respond in the format described below, and you should not output comments or other information. \\
\\
History\_summary: \\
The summary is... \\
\\
Subtask\_reasoning: \\
1. ...\\
2. ...\\
3. ...\\
4. ...\\
5. ...\\
6. ...\\
7. ...\\
8. ...\\
...\\
\\
\textbf{Subtask description:} \\
The current subtask is... \\
\bottomrule
\end{tabular}
}
\label{tab:subtask_reasoning}

\end{table}

\begin{table}[ht]
\caption{The prompt for reflecting on the task and evaluating success.}
\centering
\resizebox{0.99\linewidth}{!}{
\begin{tabular}{| p{20cm} |}
\toprule
\multicolumn{1}{|c|}{\textbf{Task Reflection and Success Evaluation}} \\
\midrule
You are a helpful AI assistant integrated with a humanoid robot body equipped to handle diverse tasks in the real world. Your advanced capabilities enable you to process textual and visual information, including computer application screenshots, and to control the robot body. Your task is to examine any inputs, interpret the context, and determine whether the last executed action has succeeded and caused the correct effect. \\
\midrule
\textbf{Overall task description:} \\
\texttt{<$task\_description$>} \\
\midrule
\textbf{\texttt{<$image\_introduction$>}} \\
\midrule
\textbf{Description of current image frame:} \\
\texttt{<$image\_description$>} \\
\midrule
\textbf{Key reason for the last action:} \\
\texttt{<$key\_reason\_of\_last\_action$>} \\
\midrule
\textbf{Last executed action with parameters used:} \\
\texttt{<$previous\_action\_call$>} \\
\midrule
\textbf{Error report for the last executed action:} \\
\texttt{<$executing\_action\_error$>} \\
\midrule
\textbf{Success Detection flag for the overall task:} \\
\texttt{<$success\_detection$>} \\
\midrule
\textbf{Valid action set in Python format to select the next action:} \\
\texttt{<$skill\_library$>} \\
\midrule
\textbf{Current and previous image are the same:} \\
\texttt{<$image\_same\_flag$>} \\
\midrule
\textbf{Semantic map:} \\
\texttt{<$semantic\_map$>} \\
\midrule
\textbf{Self Reflection Reasoning:} \\
1. What is the last executed action based on the text information above? \\
2. Make use of the information gathered from the images to decide if you have completed the current subtask. Pay special attention to the error report for the last executed action. \\
3. Think about your previous location and whether you have reached a new location (for example, moved from one hallway to another), based on the current observation and the semantic map. Consider if you are already in the target location for the last task. \\
5. Was the last executed action successful? Give reasons to this conclusion. You must refer to the following rules: \\
- If the last action executed was empty, then the previous action is deemed successful. \\
- If the action seemed to have no effect, pay attention to whether the robot position changed or if any of its hands move during the action execution process. \\
6. If the last action is not executed successfully, what was the most probable cause for it? You should give only one cause and refer to the following rules: \\
- The reasoning to chose the last action was wrong. \\
- If it is an interaction action, the most probable cause was that the action was unavailable or not activated in the current state. \\
- If there is any errors, analyze the cause based on them. \\
7. Pay attention to targets like "hallway", "exit", "doorway", "corridor", "passage", "open door", "hole in the wall", "opening", etc. They usually refers to the same target. Always use the word "hallway" for these. \\
\midrule
\textbf{Success Detection:} \\
Based on the last action, the current images, and the Success Detection flag, determine whether the overall task \texttt{<$task\_description$>} was successful. This assessment should consider the overall task's success, not just individual actions. \\
- If the last action executed was an empty list and \texttt{<$success\_detection$>} indicates the task is successful, then the overall task has a high chance of being considered a success. \\
- If the overall task was unsuccessful, specify the reason of failure and which steps are missing. \\
- If the overall task was successful, ONLY output ``SUCCESSFUL''. \\
\midrule
You should only respond in the format as described below.\\
Self Reflection Reasoning: \\
1. ... \\
2. ... \\
3. ... \\
\\
Success Detection: \\
... \\
\bottomrule
\end{tabular}
}
\label{tab:self-reflection}

\end{table}

\begin{table}[ht]
\centering
\caption{The prompt for decision-making and action execution in humanoid robot tasks.}
\resizebox{0.99\linewidth}{!}{
\begin{tabular}{| p{30cm} |}
\toprule
\multicolumn{1}{|c|}{\textbf{Action Planning}} \\
\midrule
You are a helpful AI assistant integrated with a humanoid robot body equipped to handle diverse tasks in the real world. Your advanced capabilities enable you to process textual and visual information, including computer application screenshots, and to control the robot body. By analyzing these inputs, you can understand the current context and situation of the robot. Use these insights to select the most suitable action for the robot to take next, given the current task. \\
\midrule
\textbf{Overall task description:} \\
\texttt{<$task\_description$>} \\
\midrule
\textbf{Subtask description:} \\
\texttt{<$subtask\_description$>} \\
\midrule
\textbf{Few shots:} \\
\texttt{<$few\_shots$>} \\
\midrule
\textbf{\texttt{<$image\_introduction$>}} \\
\midrule
\textbf{Description of current image:} \\
\texttt{<$image\_description$>} \\
\midrule
\textbf{Current and previous image are the same:} \\
\texttt{<$image\_same\_flag$>} \\
\midrule
\textbf{Last executed action:} \\
\texttt{<$previous\_action$>} \\
\midrule
\textbf{Key reason for the last action:} \\
\texttt{<$key\_reason\_of\_last\_action$>} \\
\midrule
\textbf{Self-reflection for the last executed action:} \\
\texttt{<$previous\_self\_reflection\_reasoning$>} \\
\midrule
\textbf{Summarization of recent history:} \\
\texttt{<$previous\_summarization$>} \\
\midrule
\textbf{Valid action set in Python format to select the next action:} \\
\texttt{<$skill\_library$>} \\
\midrule
\textbf{Success detection for overall task:} \\
\texttt{<$success\_detection$>} \\
\midrule
\textbf{Semantic map:} \\
\texttt{<$semantic\_map$>} \\
\midrule
\textbf{Decision Making Reasoning:} \\
1. Does \texttt{<$success\_detection$>} mean the overall task was successful? If successful, ignore questions 2 to 16. Otherwise, do not make conclusion before answering the other questions. \\
2. When going to a location target, in which hallway is it located? If you are already in the right hallway, DO NOT move to the wrong hallway. Make sure where you are first. Just because you can see another area, it does not mean you are in it. \\
3. If the previous action failed with "Move failed" because the target was not in the latest image, NEVER try the same action with the same target a second time! \\
4. If the target is not visible in the current image, you MUST SEARCH for the target (unless the previous action was already a search). \\
5. If you are not in the target area, and the previous action was already a search for the target, then try to find a way to move towards the target area first. For example, follow down a hallway connected to the target area. \\
6. If the current image is the same as the one from the previous iteration, DO NOT output the same action as the previous step. \\
7. Summarize the contents of recent history, mainly focusing on historical tasks and behaviors. \\
8. Summarize the contents of self-reflection for the last executed action, and do not be distracted by other information. \\
9. Think about the previous location you were in and if you have reached a new place based on the current observation. Think if you are already in the location required for the task. DO NOT rely on door labels. \\
10. Keep in mind the position of the target, even if it is no longer in the current frame. \\
11. You should search the place after any movement forward towards a target that was a hallway. \\
12. If you have to move, try to move towards some target that you can see in front of you (like the ground near your destination). \\
13. If you do not have a skill to finish the desired task, use \texttt{speak("request\_text")} to request help. \\
14. Which skill in the skill library above is the most related to how to conduct the next step of the current subtask? \\
15. This is the most critical question. Based on the action rules and self-reflection, what should be the most suitable action in the skill library for the next step? You should analyze the possible effects of the action step by step. \\

\textbf{Actions:} \\
The best action, or first action step in a short sequence of actions, to execute next towards achieving the goal. Pay attention to the names of the available skills and the previous skills already executed, if any. Pay special attention to the coordinates or direction of any action that needs them. Do not make assumptions about the location of objects or UI elements, analyze in detail any provided images. You should also pay attention to the following action rules: \\
1. If \texttt{<$success\_detection$>} means the overall task was successful or equal to "True", then output action MUST be empty like ''. Be careful to first check that the task was really successful. \\
2. You should output actions in Python code format and specify any necessary parameters to execute that action. Only use function names and argument names exactly as shown in the skill library of valid actions. If a function has parameters, you should also include their names and decide their values, like \texttt{turn\_right("small")}. If an action does not have a parameter, just output the action function call, like \texttt{go\_back()}. \\
3. You cannot open doors, so NEVER go to doors to open them. To move between rooms or areas, ALWAYS use open doorways or passages (openings) toward hallways. You should only move towards a closed door to stand in front of it if that is the final location target for the task. \\
4. Given the current situation and task, you should only choose the most suitable action from the skill library. You cannot use actions that are not in the skill library. \\
5. If you are walking down a hallway to try to find a target, you MUST perform a search for the target after any movement towards a hallway target. \\
\\
\textbf{Key reason for the last action:}  Summarize the key reasons for choosing this action to execute. \\
\\
\midrule
You should only respond in the format described below. In your reasoning for the chosen actions, also describe which object or area you decided to interact with and why. DO NOT change the title of each item in the response. You should not output other comments or information besides the format below. \\

\\

Decision\_Making\_Reasoning: \\
1. ... \\ 
2. ... \\ 
3. ... \\
4. ... \\
5. ... \\
6. ... \\
7. ... \\
8. ... \\
9. ... \\
10. ... \\
11. ... \\
12. ... \\
... \\
\\
Actions:\\
```python\\
    action(args1=x,args2=y)\\
```\\
\\
Key\_reason\_of\_last\_action:\\
...\\
\bottomrule
\end{tabular}
}
\label{tab:action_planning}
\end{table}

\begin{table}[ht]
\caption{The shorter version of the prompt we used for decision making and action execution of humanoid robot}
\centering
\resizebox{0.99\linewidth}{!}{
\begin{tabular}{| p{18cm} |}
\toprule
\textbf{Decision Making and Action Execution for Humanoid Robot} \\
\midrule
\textbf{\texttt{<$image\_introduction$>}} \\
\midrule
\textbf{Overall task description:} \\
\texttt{<$task\_description$>} \\
\midrule
\textbf{Subtask description:} \\
\texttt{<$subtask\_description$>} \\
\midrule
\textbf{Map:} \\
\texttt{<$semantic\_map$>} \\
\midrule
\textbf{Valid actions to select the next action:} \\
\texttt{<$skill\_library$>} \\
\midrule
You should only respond in the format described below.\\
\\
Action:\\
\texttt{```python}\\
    \texttt{action(args1=x,args2=y)}\\
\texttt{```}\\
\bottomrule
\end{tabular}
}
\label{tab:action_planning_short}

\end{table}

\end{document}